\documentclass{article} %
\usepackage{iclr2027_conference,times}

\usepackage{amsmath,amsfonts,bm}

\def\eqref#1{equation~\ref{#1}}

\def\1{\bm{1}}

\DeclareMathAlphabet{\mathsfit}{\encodingdefault}{\sfdefault}{m}{sl}
\SetMathAlphabet{\mathsfit}{bold}{\encodingdefault}{\sfdefault}{bx}{n}

\usepackage{hyperref}
\usepackage{url}
\usepackage{amssymb}
\usepackage{graphicx}
\usepackage{subcaption}
\usepackage{float}
\usepackage{wrapfig}
\usepackage{tabularray}
\UseTblrLibrary{booktabs}
\usepackage{rotating}
\usepackage{multirow} 
\usepackage{amsmath,amsfonts,bm}
\usepackage{tabularx}
\usepackage{pdflscape}
\usepackage{booktabs}
\usepackage{xcolor}

\usepackage{array}

\usepackage{placeins}
\usepackage{dblfloatfix}

\usepackage{pgfplots}
\pgfplotsset{compat=1.18}
\usepackage[normalem]{ulem}
\usepackage{adjustbox}  %
\usepackage{enumitem}
\usepackage{fancyvrb}
\usepackage{fvextra}
\usepackage{soul}
\usepackage{soulpos}
\usepackage[most]{tcolorbox}

\title{Beyond Equation Search: Conjecture-Guided Discovery with SCOPE and ClosureBench}

\author{Yuxiao Hu$^{1,2}$, Qian Li$^{2,3}$, Ding Wang$^{2,3}$, Siheng Chen$^{3}$, Linfeng Zhang$^{4}$,\\ \textbf{Jinyue Yan}$^{1}$, \textbf{Yuntian Chen}$^{2}$\thanks{Corresponding author} \\
$^{1}$The Hong Kong Polytechnic University, Hong Kong, China\\
$^{2}$Eastern Institute of Technology, Ningbo, China\\
$^3$Shanghai Jiao Tong University, Shanghai, China\\
$^4$DP Technology, Beijing, China\\
\texttt{yuxiao.hu@connect.polyu.hk},\texttt{qianl01205@sjtu.edu.cn}\\
\And
}

\iclrfinalcopy

\begin{document}

\maketitle
\lhead{Preprint}

\begin{abstract}
Most symbolic regression methods optimize candidate equations, but improving an equation's fit does not establish whether its underlying scientific conjecture should be retained, revised, or rejected. A conjecture may admit multiple mathematical realizations, and similarly accurate equations may imply different explanations. This motivates a shift from refining individual expressions to developing and testing the conjectures that guide them. We introduce \textbf{SCOPE}, a framework that sustains multiple conjecture-driven research trajectories. Multiple scientist agents independently pursue conjectures through equation modeling, question formulation, and active experimentation, then exchange evidence to revise their conjectures and research directions. Their findings are consolidated into a scientific report linking models, conclusions, and evidence. Existing equation-discovery benchmarks primarily assess recovered expressions, leaving the evidence and scientific conclusions established through the investigation underexamined. We therefore introduce \textbf{ClosureBench}, an interactive benchmark of 60 closure-discovery tasks across five scientific domains. ClosureBench focuses on closure-discovery tasks to give equation discovery a concrete scientific purpose. Each task embeds an unknown algebraic relation in a known PDE system, where agents control experimental conditions in OpenFOAM to acquire new data for equation discovery. Beyond equation accuracy, ClosureBench proposes three metrics to evaluate the discovery process: evidence faithfulness, mechanism insight, and experimental effectiveness. Compared with diverse baselines, SCOPE achieves higher equation accuracy on both ClosureBench and a real-world scientific discovery case, while producing more accurate, reliable, and effective scientific discoveries.
\end{abstract}

\section{Introduction}
Discovering laws from observations is a central goal of science. Equations express these laws as quantitative models that can be tested and used to make predictions, making automated equation discovery an important direction in AI for science~\citep{wang2023scientific,zhang2025artificial}. Recent advances in Symbolic Regression (SR) and Large Language Models (LLMs) have improved the generation and refinement of candidate equations by combining scientific knowledge with numerical optimization~\citep{karniadakis2021physics,krenn2022scientific}. These methods search for accurate, concise expressions and use feedback from observed data to improve their structure and parameters.

Most SR methods center optimization on candidate equations, using data feedback to refine their structure. Scientific discovery, however, often develops broader conjectures, with equations serving as mathematical realizations and tools for testing. A conjecture can admit multiple equations, while equations with similar predictive accuracy may support different explanations. The performance of a equation alone is therefore insufficient to determine whether its underlying conjecture should be retained, revised, or rejected. For example, Copernicus’s heliocentric model initially offered no predictive advantage over contemporary geocentric models. Kepler later used Tycho Brahe’s precise observations to revise orbital shapes and motion laws, advancing the heliocentric research direction. A model’s shortcomings thus need not invalidate its underlying research direction, nor does a better fit necessarily provide a more compelling scientific explanation. Automated discovery therefore calls for a loop centered on developing and testing scientific conjectures through equations: conjectures guide modeling, models reveal unresolved questions, and new experimental evidence informs revisions to conjectures, models, and research directions. 

To this end, we introduce \textbf{SCOPE} (\textbf{S}cientific \textbf{C}onjecture-\textbf{O}riented \textbf{P}arallel \textbf{E}xploration), a framework for parallel scientific agent investigation guided by structured conjectures. Each scientist agent pursues a complete research trajectory rather than merely optimizing candidate equations. After initial data analysis and literature review, each scientist formulates a structured conjecture containing a natural-language explanation, variable relationships, testable predictions, and candidate equation space. This conjecture guides symbolic regression toward accurate equations. Each scientist then uses elite equations and existing evidence to identify an unresolved question, conduct targeted analysis, and design executable experimental conditions. New observations from experiments inform conjecture revision and the next round of modeling and investigation. Every two rounds, the scientists hold a roundtable discussion to exchange conjecture, elite equations, experimental evidence, and open questions. They then independently adjust their research directions. SCOPE stops when the budget is exhausted or the scientists reach consensus during a roundtable discussion. In the loop of SCOPE, accurate equations serve as both valuable outcomes and quantitative tools for testing conjectures. The final scientific report integrates models, supported conclusions, and open questions, documenting what has been learned and the evidence supporting it.

Evaluating this research process requires agents to actively acquire evidence through experiments and assessment of the full discovery process—not just equation recovery, but also the scientific findings established and the evidence supporting them. We therefore introduce \textbf{ClosureBench}, an interactive benchmark of 60 closure-discovery tasks across five scientific domains. Each task embeds an unknown algebraic closure relation in a known PDE system, giving the target equation a concrete scientific roles. In many scientific experiments, scientists control experimental conditions rather than directly select local states to observe. ClosureBench captures this constraint through OpenFOAM: agents specify initial, boundary, or operating conditions, and OpenFOAM produce the resulting states and measurements. Experiment design therefore requires translating scientific questions into experimental conditions that can generate the evidence needed to answer it. Beyond equation accuracy, ClosureBench introduces three new metrics to evaluate conjecture-centered automated scientific discovery: Evidence Faithfulness (EF), Mechanism Insight (MI), and Experimental Effectiveness (EE). We evaluate SCOPE on ClosureBench and real-world scientific discovery tasks, where it improves both equation recovery and generalization while producing more accurate, reliable, and effective scientific reports.

Our contributions come in three areas: (1)We frame scientific discovery as the development of interacting, conjecture-driven research trajectories that connect equation modeling, experimentation, and evidence-based revision. (2) We introduce SCOPE, a multi-scientist framework combining independent investigation with periodic evidence exchange to produce scientific reports grounded in experimental findings. (3) We introduce ClosureBench for discovering hidden closure relations in PDE systems through OpenFOAM experiments. Beyond recovered equations, ClosureBench evaluates the quality of scientific discovery using three new metrics.

\begin{figure*}
 \includegraphics[width=\textwidth]{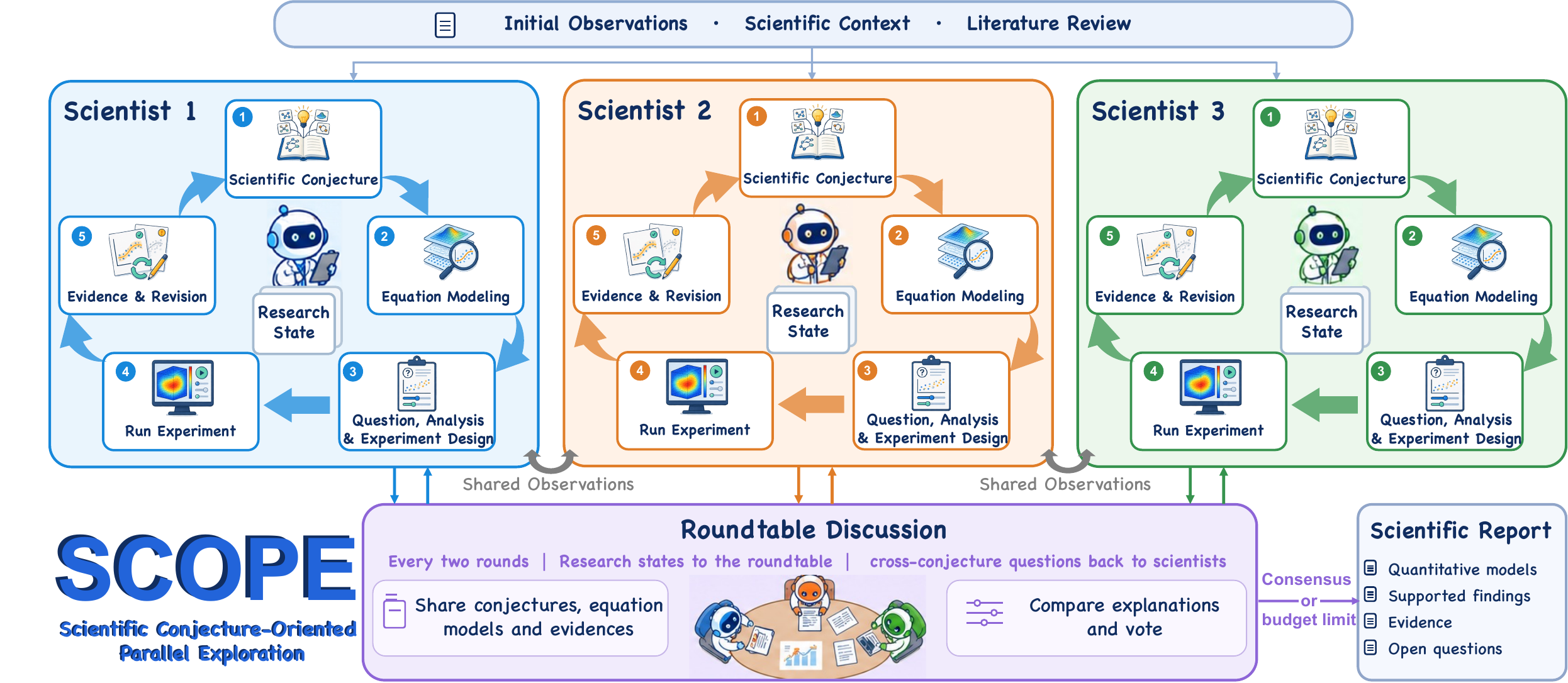}
  \caption{Overview of SCOPE. SCOPE organizes scientific discovery around parallel, conjecture-driven investigations. Scientists use equation modeling and active experiments to test and revise conjectures, while roundtable discussions exchange evidence and guide further exploration. The final report connects quantitative models with supported scientific findings and open questions.}
  \label{overview}
\end{figure*}

\section{Related Work}
\textbf{Symbolic regression algorithms.}
Symbolic regression (SR) seeks explicit expressions that balance predictive accuracy and simplicity~\citep{scholl2025parfam,merler2024context,wang2026llm}. Search-based methods include genetic programming~\citep{gplearn}, Monte Carlo tree search~\citep{spl}, and parallel multi-population evolution in PySR~\citep{pysr}. Pretrained methods learn equation priors from synthetic data~\citep{meidani2023snip,e2e,gensr,transr,shojaee2023transformer,zhang2025rag}. LLM-based approaches use scientific knowledge and program generation to guide search~\citep{llmsr,lasr,zhang2026llm,lirobust,guo2025sr}, while SR-Scientist~\citep{xia2025sr} integrates data analysis, tool use, and iterative equation refinement. LLM-AutoSciLab and LLM-ACES further incorporate active experimentation to distinguish and refine candidate mechanisms~\citep{kabra2026llm,abhyankar2026llm}. However, in their equation-discovery settings, the loop remains organized around candidate equations and their recovery. 

SCOPE instead organizes the loop around the development and testing of scientific conjectures, with equation recovery as one stage. The goal is to establish scientific understanding supported by accurate models and traceable evidence, with the findings and remaining uncertainties integrated into a scientific report.

\textbf{Symbolic regression benchmarks.}
Standard SR benchmarks primarily evaluate equation recovery from fixed observations. Feynman~\citep{udrescu2020ai} focuses on physical equations, SRBench~\citep{la2021contemporary} covers synthetic and real-world tasks, and LLM-SRBench~\citep{llmsrbench} introduces memorization-resistant equations. However, fixed datasets cannot evaluate how agents design experiments to obtain evidence needed for discovery. Interactive benchmarks address this limitation: NewtonBench~\citep{zheng2025newtonbench} enables agents to probe simulated systems and recover hidden counterfactual laws, while ActiveSciBench~\citep{kabra2026llm} provides budgeted experiments for enzyme-kinetic equation recovery and regulatory graph discovery. Although both evaluate active exploration, their primary success criteria remain the recovery of target equations or structures. These criteria do not directly assess which scientific conclusions agents establish or whether their experimental evidence supports those conclusions.

ClosureBench extends evaluation beyond equation recovery to the scientific discovery process, including active experimentation and the quality of scientific findings. It embeds unknown algebraic closures in known PDE systems, grounding each investigation in a concrete physical problem. Agents specify experimental conditions through OpenFOAM to obtain new observations. Evaluation covers equation accuracy and generalization, scientific property identification, evidential support for conclusions, and progress enabled by experiments.

\section{SCOPE: Scientific Conjecture-Oriented Parallel Exploration}\label{SCOPE}

SCOPE maintains \(M\) scientist agents and takes initial observations \(\mathcal D_0\), variable descriptions, and scientific context \(\mathcal C\), with access to literature retrieval, data analysis, symbolic regression, and experimental tools. After round \(r\), all scientists share the cumulative observations \(\mathcal D_r\), while scientist \(i\) maintains its own research state:
$$
S_i^r=(H_i^r,\mathcal V_i^r,\mathcal Q_i^r,\mathcal L_i^r),
$$
where \(H_i^r\) is its current conjecture, \(\mathcal V_i^r\) contains accessible candidate equations, \(\mathcal Q_i^r\) tracks research questions and their status, \(\mathcal L_i^r\) records analyses, experiments, and discussions. The initial state is
\(
S_i^0=\left(H_i^0,\varnothing,\varnothing,\mathcal L_i^0\right)\). \(\mathcal L_i^0\) records the initial data analysis and literature findings used to form \(H_i^0\).
\begin{table}[H]
\centering
\caption{Comparison of scientific discovery frameworks across key design dimensions.}
\label{tab:framework_comparison}

\begingroup
\fontsize{6.5}{7.5}\selectfont
\setlength{\tabcolsep}{2.5pt}
\renewcommand{\arraystretch}{1.1}

\def\scopeTableCell#1{%
  \begingroup
  \renewcommand{\arraystretch}{0.9}%
  \begin{tabular}[c]{@{}c@{}}#1\end{tabular}%
  \endgroup
}

\def\scopeTableRefs#1{%
  {\fontsize{5.5}{6}\selectfont\citep{#1}}%
}

\begin{tabular*}{\textwidth}{
  @{\extracolsep{\fill}}cccccc@{}
}
\toprule

\scopeTableCell{\textbf{Method / Framework}}
&
\scopeTableCell{\textbf{Primary object}\\
                \textbf{of refinement}}
&
\scopeTableCell{\textbf{Symbolic}\\
                \textbf{equation modeling}}
&
\scopeTableCell{\textbf{Active}\\
                \textbf{experimentation}}
&
\scopeTableCell{\textbf{Parallel}\\
                \textbf{research streams}}
&
\scopeTableCell{\textbf{Primary}\\
                \textbf{output}}
\\
\midrule

\scopeTableCell{Equation search\\
  \scopeTableRefs{pysr,llmsr}}
&
\scopeTableCell{Equations}
&
\scopeTableCell{$\checkmark$}
&
\scopeTableCell{---}
&
\scopeTableCell{---}
&
\scopeTableCell{Equations}
\\
\addlinespace[1.5pt]

\scopeTableCell{Agentic equation discovery\\
  \scopeTableRefs{xia2025sr}}
&
\scopeTableCell{Equations}
&
\scopeTableCell{$\checkmark$}
&
\scopeTableCell{---}
&
\scopeTableCell{---}
&
\scopeTableCell{Equations}
\\
\addlinespace[1.5pt]

\scopeTableCell{Active structure discovery\\
  \scopeTableRefs{kabra2026llm,abhyankar2026llm}}
&
\scopeTableCell{Equations /\\network structures}
&
\scopeTableCell{$\checkmark$}
&
\scopeTableCell{$\checkmark$}
&
\scopeTableCell{---}
&
\scopeTableCell{Equations /\\network structures}
\\
\addlinespace[1.5pt]

\scopeTableCell{Scientific research agents\\
  \scopeTableRefs{gottweis2026accelerating,mitchener2025kosmos}}
&
\scopeTableCell{Hypotheses /\\research plans}
&
\scopeTableCell{---}
&
\scopeTableCell{---}
&
\scopeTableCell{---}
&
\scopeTableCell{Research proposals}
\\

\midrule

\scopeTableCell{\textbf{SCOPE (Ours)}}
&
\scopeTableCell{\textbf{Conjecture-driven}\\
                \textbf{research trajectories}}
&
\scopeTableCell{$\checkmark$}
&
\scopeTableCell{$\checkmark$}
&
\scopeTableCell{$\checkmark$}
&
\scopeTableCell{\textbf{Equations +}\\
                \textbf{scientific report}}
\\

\bottomrule
\end{tabular*}

\vspace{3pt}
\begin{minipage}{\textwidth}
\fontsize{6}{7}\selectfont
\textit{Symbolic equation modeling}: explicit search over equation
structures and parameters.
\textit{Active experimentation}: executing selected experiments
and incorporating new observations.
\textit{Parallel research streams}: persistent branches with
their own conjectures, questions, and investigation histories.
$\checkmark$: explicitly described; ---: not explicitly specified.
\end{minipage}

\endgroup
\end{table}

\subsection{Conjecture-Guided Equation Modeling}

\textbf{Structured conjectures.} Initially, scientist $i$ formulates conjectures from data analysis and literature review \(H_i^0=\operatorname{Init}_{\mathrm{LLM}}
\left(\mathcal D_0,\mathcal C,\mathcal L_i^0\right)\). At the beginning of round \(r\), each conjecture is represented as
$$
H_i^r=(m_i^r,\mathbf v_i^r,\mathcal R_i^r,\mathcal P_i^r,\mathcal F_i^r,e_i^r),
$$
comprising a mechanistic explanation \(m_i^r\), relevant variables $\mathbf v_i^r$, qualitative relationships $\mathcal R_i^r$, testable predictions $\mathcal P_i^r$, candidate equation families $\mathcal F_i^r$, and an evidence summary $e_i^r$. For example, a scientist agent may conjecture that finite active-site capacity limits the reaction rate \(y\) as reactant concentration \(x\) increases. This implies an increasing, saturating relationship and predicts that doubling \(x\) produces a smaller relative increase in \(y\) at high concentrations. Candidate realizations may include $y=\frac{ax}{b+x}$ or $y=a(1-e^{-bx})$, $a,b>0,\;x\geq0$. The evidence summary might record diminishing rate increases while noting that neither a plateau nor site occupancy has been observed. A conjecture thus guides a family of mathematical realizations rather than specifying a single equation, while making its predictions and evidential limits explicit.

\textbf{Equation Subspace Expansion.} Given \(H_i^r\), the LLM refines, mutates, and recombines expressions in \(\mathcal F_i^r\) to construct a conjecture-guided prior subspace, recording each candidate’s origin and connection to the conjecture. The expanded set \(\widetilde{\mathcal F}_i^r\) provides prior guidance for symbolic regression.

\textbf{Guided symbolic regression.} SCOPE supports PySR~\citep{pysr} and LLM-SR~\citep{llmsr} through a common interface. \(\widetilde{\mathcal F}_i^r\) initialize part of the evolutionary population in PySR or serve as initial executable programs in LLM-SR. Previously discovered candidates in \(\mathcal V_i^r\) can also seed subsequent rounds. Both SR solvers optimize expression structure and parameters on \(\mathcal D_r\), producing quantitative models for further investigation.

\textbf{Elite selection.} Let \(\mathcal G_i^r\) denote the equations returned by symbolic regression tools. SCOPE combines them with previously accessible candidates: \(\widetilde{\mathcal V}_i^r=\operatorname{Filter}\left(\mathcal V_i^r\cup\mathcal G_i^r;\mathcal D_r\right)\), where \(\operatorname{Filter}\) removes duplicate and non-executable expressions and candidates with non-finite predictions. All retained candidates are evaluated on \(\mathcal D_r\). SCOPE then selects at most \(K\) elite equations:\(\mathcal E_i^r=\operatorname{Select}_K\bigl(\operatorname{Pareto}_{f\in\widetilde{\mathcal V}_i^r}\bigl(\ell(f;\mathcal D_r),c(f)\bigr)\bigr),\)where \(\ell\) is prediction error and \(c\) measures expression complexity. Selection preserves the lowest-error and simplest expressions while favoring diverse variable-dependency patterns. The elite equations and their evaluation results support subsequent question formulation and conjecture testing.

\subsection{Question-Driven Investigation}
\paragraph{Research question formulation.}
Scientist \(i\) reviews the current conjecture \(H_i^r\), elite equations \(\mathcal E_i^r\), research records \(\mathcal L_i^r\), and open questions in \(\mathcal Q_i^r\), including those raised during previous roundtables. SCOPE constructs a question context, which contains three diagnostics: predictive disagreements among elite equations on \(\mathcal D_r\), recurring residual patterns across elite equations, and predictions or variable ranges not yet supported by evidence. The LLM selects an existing question or proposes a new one whose resolution could inform conjecture revision through data analysis or a feasible experiment. The new proposed question and its supporting references are recorded in \(\mathcal Q_i^r\) to guide subsequent investigation.

\textbf{Targeted analysis and experimentation.} Within a fixed tool budget, each scientist investigates the selected question through three tools: (1) \texttt{data analyzer}: diagnosing residuals and input coverage; (2) \texttt{existing-row contrast}: examining response or residual trends within experimental conditions and aggregating them across conditions; (3) \texttt{equation-property check}: checking candidate dependencies, interactions, local behavior, parameter stability, and identifiability. The scientist then designs an experiment to address the remaining gap or to test the conclusion drawn from existing evidence.

In physical experiments, scientists often control experimental conditions rather than directly select the observed states. Each scientist agent therefore identifies informative observations by the three tools above, then translates these requirements into executable conditions. An experimental request specifies conditions \(a\), required measurements, expected outcomes, and falsification criteria, with feasibility checked before execution. The latter two guide assessment of the returned observations.

\textbf{Evidence assessment and conjecture revision.} Let \(\mathcal A_r\) denote the experiments completed in round \(r\), with \(\mathcal O(a)\) returning observations under conditions \(a\). The shared dataset is updated as
$$
\mathcal D_{r+1}=\mathcal D_r\cup\bigcup_{a\in\mathcal A_r}\mathcal O(a).
$$
Each scientist analyzes the new observations, reassesses model residuals, and tests the predictions against the expected outcomes and falsification criteria specified before execution. The program applies these criteria to classify each prediction as supported, refuted, or inconclusive. Failure to obtain the required observations yields an inconclusive result. Let \(\mathcal Z_i^r\) collect scientist \(i\)’s analysis and experimental assessment results. The conjecture is updated as
$$
H_i^{r+1}=\operatorname{Update}_{\mathrm{LLM}}\left(H_i^r,\mathcal E_i^r,\mathcal Z_i^r\right).
$$
The LLM compares this evidence with the current conjecture and elite equations, identifies supported, challenged, and unresolved claims, and revises the mechanistic explanation \(m_i^r\), relevant variables $\mathbf v_i^r$, qualitative relationships $\mathcal R_i^r$, predictions $\mathcal P_i^r$, or candidate equation families $\mathcal F_i^r$ accordingly. Each revision cites its supporting evidence, and the evidence summary is updated accordingly. \(\mathcal Q_i^{r+1}\) retains question histories, including the new question, updated statuses, and evidence references, while \(\mathcal L_i^{r+1}\) records analyses, experimental assessments, and conjecture revisions. Unresolved questions guide further investigation. The updated conjecture \(H_i^{r+1}\) and observations \(\mathcal D_{r+1}\) jointly guide the next modeling round.

\subsection{Roundtable Discussion and Scientific Reporting}\label{3.3}
\textbf{Roundtable discussion and voting.} SCOPE holds a roundtable every two rounds, after experimental assessment and conjecture revision. Each scientist shares its revised conjecture \(H_i^{r+1}\), elite equations \(\mathcal E_i^r\), question and evidence records. Scientists assess all trajectories, including their own, by comparing scientific explanations, supporting evidence, and prediction errors of elite equations \(\mathcal E_i^r\) on \(\mathcal D_{r+1}\). Each then independently votes for the most convincing trajectory and provides a rationale. Unanimous selection triggers early termination; otherwise, exploration continues for up to ten rounds, comprising five roundtables.

\textbf{Research after exchange.} When consensus is not reached and rounds remain, each scientist formulates a new question from other trajectories’ conjectures, models, and evidence. The question specifies the claim to be tested, its relation to the scientist’s own conjecture, the observations needed, and how different outcomes would affect that conjecture. The question enters \(\mathcal Q_i^{r+1}\) for consideration alongside questions arising within the trajectory, while discussion and voting records enter \(\mathcal L_i^{r+1}\). Cross-trajectory findings thus inform subsequent analysis and experimentation, allowing new evidence to drive conjecture revision. These additions complete the \(S_i^{r+1}\) used in the next round.

\textbf{Final conclusions and scientific reporting.} 
Upon unanimous agreement, SCOPE selects the consensus trajectory. Otherwise, it selects the trajectory receiving the most votes at the fifth roundtable, breaking ties by lower mean prediction error. Within the selected trajectory, the elite equation with the lowest prediction error on \(\mathcal D_{\mathrm{final}}\) is returned as the final equation. Other trajectories are retained as alternative explanations.

SCOPE provides the LLM with the task description, selected research trajectory and final equation, and an evidence summary compiled during the investigation to generate a scientific report. We additionally provide a reference report for in-context guidance~\citep{yang2026multimodal}, together with writing guidelines for organizing findings, connecting evidence with interpretations, and constructing a coherent scientific narrative. The report's scientific statements are grounded in evidence from the current investigation.

\section{ClosureBench}
\label{sec:benchmark}
Evaluating scientific discovery requires assessing not only the derived equations, but also the discovery process: how evidence is acquired and whether it supports the agent’s conclusions. We introduce \textbf{ClosureBench}, an interactive benchmark containing 60 closure-discovery tasks across five scientific domains. Each task embeds an unknown algebraic relation in a known PDE system. An OpenFOAM interface supports experimental evidence acquisition, while evaluation measures equation accuracy, identified scientific properties, evidential support, and improvement through interactive research.

\subsection{Closure-Discovery Tasks}
Closure relations describe physical processes left unspecified by known governing equations. Discovering these relations from data is an important problem in scientific modeling~\citep{sanderse2024scientific}. ClosureBench uses this setting to give equation discovery a concrete scientific purpose.

\textbf{Task formulation.}
Each task consists of a known PDE system with an unknown algebraic closure:
\[
\mathcal R(\mathbf U,\mathbf a,c_\star)=0,
\qquad
c_\star
=
g_\star\!\left(\mathbf z\right).
\]
Here, $\mathbf U$ denotes the spatiotemporal physical fields, and $\mathbf a$ specifies controllable initial, boundary, or operating conditions, such as initial temperature, inlet velocity, or external loading. The known operator $\mathcal R$ defines the governing equations and how the closure quantity enters them. The unknown target expression $g_\star$ maps local physical inputs $\mathbf z$ to the closure quantity $c_\star$. These inputs $\mathbf z$ comprise quantities such as temperature and velocity. For example, an advection--diffusion energy equation describes temperature transport and may include an interphase heat-exchange source specified by an unknown closure. In this case, $\mathbf z$ may contain the two phase temperatures and local flow velocity, while $g_\star(\mathbf z)$ determines the corresponding heat-exchange rate.

\textbf{Task construction.}
We first select classical PDE systems and specify the process, variables, and coupling associated with each unknown closure. Inspired by LSR-Synth's use of scientific context to construct synthetic discovery tasks~\citep{llmsrbench}, we use LLMs to generates and structurally modifies  algebraic closures for each PDE system. Researchers review dimensional consistency, physical constraints, and limiting behavior. Candidates are then embedded in OpenFOAM and checked for numerical stability, conservation consistency, and mesh and time-step sensitivity. Each task exposes its scientific context, variable names, units, physical meanings, and experimental interface, but not its hidden closure.

\subsection{Condition-Driven Interactive Experiments}
\textbf{Experimental setting and observation generation.}
In complex physical systems, researchers can set experimental conditions but cannot directly prescribe the local states they observe. ClosureBench preserves this constraint through
\[
\mathbf a
\xrightarrow{\mathrm{OpenFOAM}}
\mathbf U
\longrightarrow
\mathcal D(\mathbf a).
\]
Agents propose experimental conditions $\mathbf a$ including initial, boundary, or operating conditions. OpenFOAM solves the corresponding PDE, and observations $\mathcal D(\mathbf a)=\{(\mathbf z_j, c_j)\}_{j=1}^{n_{\mathbf a}}$ are extracted from the resulting fields. $n_{\mathbf a}$ is the observation count, $\mathbf z_j$ is extracted from the simulated fields, and $c_j$ denotes the value of the target expression \(g\) at \(\mathbf z_j\). Agents must therefore determine which conditions can produce evidence that addresses their scientific questions.

\textbf{Data budget.}
All methods share the same 1,000 initial observations. SR methods with fixed-data receive 3,000 additional observations generated under different experimental conditions randomly sampled in advance from the public control ranges. Interactive methods instead select conditions during research. For each task, an interactive method may make at most \(B_{\max}=30\) additional simulation calls and receive up to 3,000 additional observations; SCOPE's scientists share both limits. The common limit of 4,000 observations enables comparison between random and question-driven conditions under the same maximum data budget.

\subsection{Evaluation Protocol}\label{4.3}
Beyond the equation-accuracy metrics NMSE, Acc@1\% and Acc@0.1\% adopted in prior work~\citep{e2e,xia2025sr}, we introduce three additional metrics to evaluate the reliability, accuracy, and effectiveness of the scientific research records. The single-task equation-metric definitions are given in Appendix~\ref{app:equation-metrics}.

\textbf{Evidence Faithfulness (EF)} measures whether verifiable report conclusions are supported by the evidence collected during the investigation. A fixed LLM evaluator extracts distinct statements and checks their direction, magnitude, scope, and strength of conclusion against the research record. For $N$ deduplicated statements, let $s_j\in\{0,1\}$ indicate whether statement $j$ is fully supported:
\[
\mathrm{EF}
=
\frac{1}{N}\sum_{j=1}^{N}s_j.
\]

\textbf{Mechanism Insight (MI)} measures whether the report identifies scientific properties of the hidden relation. Each task has a predefined, researcher-reviewed reference set $\mathcal F_{\mathrm{ref}}$, covering variable dependence, trends, interactions, and limiting behavior. The evaluator extracts report claims $\mathcal C_{\mathrm{report}}$ and scores \(1\), \(0.5\), or \(0\) for full, partial, or no valid correspondence. We match report claims to reference properties one-to-one. Let \(W\) denote the maximum total weight over all valid matchings:
\[
\mathrm{MI}
=
\frac{2W}
{|\mathcal C_{\mathrm{report}}|+|\mathcal F_{\mathrm{ref}}|}.
\]

\textbf{Experimental Effectiveness (EE)} measures the improvement in equation accuracy over the research process. For task \(i\), let \(A_{s,i}^{(0)}\) and \(A_{s,i}^{(T)}\) be the fractions of test points on split \(s\in\{\mathrm{ID},\mathrm{OOD}\}\) whose relative error is at most 1\% for the initial and final equations. We define
\[
\mathrm{EE}_i
=\frac{
  \left(A_{\mathrm{ID},i}^{(T)}-A_{\mathrm{ID},i}^{(0)}\right)
 +\left(A_{\mathrm{OOD},i}^{(T)}-A_{\mathrm{OOD},i}^{(0)}\right)
}{2}.
\]
The initial equation uses only the initial observations, while the final equation is the final recovered equation; both are evaluated on the hidden test sets after research concludes. Across tasks, we take the average over valid initial--final evaluation pairs and report the results in percentage points.

\textbf{Evaluation reliability.}
For EF and MI, a fixed, method-blind LLM evaluator extracts statements and makes judgments, while the scores are computed programmatically. The human-validation protocol uses reports sampled across domains and methods, independently annotated by two domain-informed researchers blinded to method identities and LLM judgments. Evaluator prompts, edge cases, and human–human/LLM–human agreement are provided in the appendix.

\section{Experiments}
\label{sec:experiments}

We evaluate SCOPE from two perspectives: final-equation accuracy and discovery quality. We report ID and OOD accuracy for the final equations of all methods. For agents capable of actively conducting computational experiments, we additionally assess the quality of experimental decisions, evidence utilization, and mechanistic explanations. All tasks, splits, budgets, and metrics follow the unified protocol in Section~\ref{sec:benchmark}.

Additional results, details and analyses---including candidate-set capacity sensitivity, human validation of the LLM evaluators, representative ClosureBench tasks across five scientific domains, a real-world scientific discovery case, and complete annotated reports for the low-induced crystallization case---are provided in the appendix.

\subsection{Baselines and Implementation Details}
\label{sec:baselines}

\paragraph{Fixed-data symbolic regression.}
We compare against four conventional symbolic regression methods---E2E, DSR, uDSR, and PySR---spanning pretrained expression generation, reinforcement learning, and evolutionary symbolic search. We further compare against three LLM-based SR methods: LaSR, LLM-SR, and SR-Scientist. LaSR and LLM-SR employ language models to guide equation search, whereas SR-Scientist analyzes data, constructs equations, and refines candidates from execution feedback through long-horizon tool invocation. We evaluate LLM-based methods with Qwen3.6-35B-A3B, GPT-5.4-mini, and Kimi K2.6.
These methods do not conduct interactive experiments to obtain new data.

\paragraph{Interactive scientific agents.}
We adopt ReAct and Reflexion-Exp as representative baselines for interactive scientific discovery. ReAct~\cite{yao2022react} interleaves reasoning and acting: at each step, it reasons over the current evidence, selects a tool action, and incorporates the returned result into subsequent decisions. Reflexion-Exp adapts Reflexion~\cite{shinn2023reflexion} to multi-round computational experimentation. Periodically, it summarizes what was supported, what failed, and what remains uncertain, stores this reflection in cross-round memory, and uses the accumulated experience to guide the next action. Unlike ReAct, Reflexion-Exp explicitly retains cross-round experience, but still develops its discovery state serially along a single trajectory. All three interactive agents have access to core scientific capabilities such as data analysis, literature retrieval, base SR methods (PySR and LLM-SR), equation validation, and computational experiments. Thus, the comparison focuses on how the discovery process is organized: stepwise reasoning and action in ReAct, sequential reflection in Reflexion-Exp, and parallel conjecture-guided investigation linking scientific questions, targeted experiments, evidence-based revision, and periodic exchange in SCOPE.

For all methods using LLMs, we impose a ceiling of 1{,}000 LLM calls per task, following the long-horizon budget used by SR-Scientist~\cite{xia2025sr}. For further details on implementation and baselines, see Appendix~\ref{app:implementation-details}. All main experiments are repeated with three independent random seeds.

\subsection{Main Results on ClosureBench}
\label{sec:main-results}

As shown in Table~\ref{tab:main_results}, SCOPE generally outperforms the fixed-data and interactive baselines in final-equation accuracy with each of the three backbones. It also achieves higher EF, MI, and EE than ReAct and Reflexion-Exp using the same backbone. These joint gains may reflect SCOPE's conjecture-centered design: active experiments extend the evidence beyond fixed observations, while parallel research trajectories exchange findings and revise competing explanations. This process may support both more accurate equations and better-grounded scientific reports.

\begin{table*}[!t]
  \centering
  \caption{Overall results on ClosureBench. EF, MI, and EE denote Evidence Faithfulness, Mechanism Insight, and Experimental Effectiveness. N/A denotes metrics not evaluated for fixed-data methods.}
  \label{tab:main_results}
  \small
  \newcommand{\scinmse}[2]{\ensuremath{#1\mathrm{e}{#2}}}
  \newcommand{\bscinmse}[2]{\ensuremath{\bm{#1\mathrm{e}{#2}}}}
  \renewcommand{\arraystretch}{0.96}
  \setlength{\tabcolsep}{3.2pt}
  \setlength{\aboverulesep}{0.35ex}
  \setlength{\belowrulesep}{0.35ex}
  \begin{adjustbox}{max width=\textwidth}
  \begin{tabular}{l>{\centering\arraybackslash}m{2.0cm}cccccc*{3}{>{\centering\arraybackslash}m{1.05cm}}}
    \toprule
    & & \multicolumn{3}{c}{ID} & \multicolumn{3}{c}{OOD} & \multicolumn{3}{c}{Discovery Quality} \\
    \cmidrule(lr){3-5}\cmidrule(lr){6-8}\cmidrule(lr){9-11}
    Method & Backbone & NMSE $\downarrow$ & Acc@1\% $\uparrow$ & Acc@0.1\% $\uparrow$
      & NMSE $\downarrow$ & Acc@1\% $\uparrow$ & Acc@0.1\% $\uparrow$
      & EF $\uparrow$ & MI $\uparrow$ & EE $\uparrow$ \\
    \midrule

    \multicolumn{11}{c}{\textbf{Fixed-data non-LLM SR}} \\
    \midrule
    E2E & -- & \scinmse{7.20}{-2} & 1.11 & 1.11 & \scinmse{2.24}{-1} & 0.00 & 0.00 & \textit{N/A} & \textit{N/A} & \textit{N/A} \\
    DSR & -- & \scinmse{1.94}{-2} & 2.22 & 1.11 & \scinmse{8.43}{-2} & 1.11 & 0.00 & \textit{N/A} & \textit{N/A} & \textit{N/A} \\
    uDSR & -- & \scinmse{2.36}{-3} & 7.78 & 2.22 & \scinmse{2.11}{-2} & 6.67 & 1.11 & \textit{N/A} & \textit{N/A} & \textit{N/A} \\
    PySR & -- & \scinmse{6.96}{-4} & 14.44 & 11.67 & \scinmse{6.99}{-3} & 12.22 & 9.44 & \textit{N/A} & \textit{N/A} & \textit{N/A} \\
    \midrule

    \multicolumn{11}{c}{\textbf{Fixed-data LLM-based SR}} \\
    \midrule
    \multirow{3}{*}{LaSR} & Qwen3.6-35B & \scinmse{2.26}{-2} & 10.56 & 7.78 & \scinmse{7.00}{-2} & 8.33 & 3.33 & \textit{N/A} & \textit{N/A} & \textit{N/A} \\
     & GPT-5.4-mini & \scinmse{4.15}{-2} & 11.11 & 10.56 & \scinmse{9.27}{-2} & 9.44 & 7.22 & \textit{N/A} & \textit{N/A} & \textit{N/A} \\
     & Kimi K2.6 & \scinmse{1.83}{-2} & 17.78 & 8.33 & \scinmse{6.37}{-2} & 8.89 & 4.44 & \textit{N/A} & \textit{N/A} & \textit{N/A} \\
    \cmidrule(lr){2-11}
    \multirow{3}{*}{LLM-SR} & Qwen3.6-35B & \scinmse{5.70}{-2} & 21.11 & 13.89 & \scinmse{2.25}{-1} & 15.56 & 13.89 & \textit{N/A} & \textit{N/A} & \textit{N/A} \\
     & GPT-5.4-mini & \scinmse{3.13}{-2} & 16.67 & 15.56 & \scinmse{6.98}{-2} & 12.78 & 11.67 & \textit{N/A} & \textit{N/A} & \textit{N/A} \\
     & Kimi K2.6 & \scinmse{1.41}{-2} & 20.00 & 13.33 & \scinmse{3.52}{-2} & 18.33 & 10.56 & \textit{N/A} & \textit{N/A} & \textit{N/A} \\
    \cmidrule(lr){2-11}
    \multirow{3}{*}{SR-Scientist} & Qwen3.6-35B & \scinmse{5.21}{-2} & 26.11 & 18.33 & \scinmse{1.51}{-2} & 18.89 & 14.44 & \textit{N/A} & \textit{N/A} & \textit{N/A} \\
     & GPT-5.4-mini & \scinmse{2.13}{-2} & 22.22 & 16.67 & \scinmse{4.91}{-2} & 16.67 & 14.44 & \textit{N/A} & \textit{N/A} & \textit{N/A} \\
     & Kimi K2.6 & \scinmse{9.69}{-3} & 24.44 & 13.33 & \scinmse{3.60}{-2} & 22.22 & 11.67 & \textit{N/A} & \textit{N/A} & \textit{N/A} \\
    \midrule

    \multicolumn{11}{c}{\textbf{Interactive agents}} \\
    \midrule
    \multirow{3}{*}{ReAct} & Qwen3.6-35B & \scinmse{5.61}{-3} & 26.67 & 17.22 & \scinmse{2.54}{-2} & 19.44 & 12.22 & 0.48 & 0.23 & 16.7 \\
     & GPT-5.4-mini & \scinmse{2.05}{-2} & 23.89 & 15.56 & \scinmse{4.20}{-2} & 20.56 & 12.78 & 0.55 & 0.20 & 13.5 \\
     & Kimi K2.6 & \scinmse{7.12}{-3} & 28.33 & 12.78 & \scinmse{3.10}{-2} & 23.33 & 11.11 & 0.56 & 0.27 & 16.4 \\
    \cmidrule(lr){2-11}
    \multirow{3}{*}{Reflexion-Exp} & Qwen3.6-35B & \scinmse{9.39}{-3} & 30.56 & 16.11 & \scinmse{2.67}{-2} & 17.78 & 13.89 & 0.50 & 0.25 & 14.5 \\
     & GPT-5.4-mini & \scinmse{9.12}{-3} & 27.22 & 17.78 & \scinmse{2.39}{-2} & 22.22 & 16.11 & 0.51 & 0.26 & 11.6 \\
     & Kimi K2.6 & \scinmse{4.12}{-3} & 32.78 & 18.89 & \scinmse{1.55}{-2} & 28.33 & 18.33 & 0.61 & 0.32 & 15.8 \\
    \cmidrule(lr){2-11}
    \multirow{3}{*}{SCOPE} & Qwen3.6-35B & \scinmse{8.95}{-4} & \textbf{41.67} & 23.89 & \scinmse{2.16}{-3} & 25.00 & 17.22 & 0.57 & 0.38 & 17.4 \\
     & GPT-5.4-mini & \bscinmse{3.18}{-4} & 33.89 & 24.44 & \bscinmse{2.47}{-4} & 32.78 & 23.33 & \textbf{0.68} & 0.37 & \textbf{19.6} \\
     & Kimi K2.6 & \scinmse{4.15}{-4} & 39.44 & \textbf{29.44} & \scinmse{1.29}{-3} & \textbf{38.33} & \textbf{27.22} & 0.66 & \textbf{0.47} & 18.9 \\
    \bottomrule
  \end{tabular}
  \end{adjustbox}
\end{table*}

\subsection{Effectiveness of Multi-Round Experiments}
\label{sec:rounds}

The left panel of Figure~\ref{fig:acquisition_dynamics} shows ID Acc@1\% over cumulative additional simulation calls for ReAct, Reflexion-Exp, and SCOPE.
SCOPE improves more rapidly in the early and middle stages, with smaller gains near the end of the budget. After 12 additional simulation calls, its ID Acc@1\% reaches 33.33\%, exceeding the final values of ReAct (26.67\%) and Reflexion-Exp (30.56\%). Both baselines also improve overall, while SCOPE finishes at 41.67\%. SCOPE's faster improvement may stem from its question-driven investigation loop, which links unresolved scientific questions to targeted experiments and revises conjectures as new evidence arrives.

\begin{figure*}[t]
  \centering
  \definecolor{scopePlotBg}{HTML}{F5F6F7}
  \definecolor{scopeGrid}{HTML}{E1E4E7}
  \definecolor{scopeAxis}{HTML}{9AA0A6}
  \definecolor{scopeBlue}{HTML}{4F6D8A}
  \definecolor{scopeReflexion}{HTML}{B9785B}
  \definecolor{scopeReact}{HTML}{718C7A}
  \definecolor{scopeSingle}{HTML}{8A7896}

  \begin{minipage}[t]{0.485\textwidth}
    \centering
    \begin{tikzpicture}
      \begin{axis}[
        width=0.98\linewidth,
        height=0.64\linewidth,
        xmin=0, xmax=30,
        ymin=10, ymax=44,
        xtick={0,6,12,18,24,30},
        ytick={10,15,20,25,30,35,40},
        xlabel={Number of simulation experiments},
        ylabel={ID Acc@1\% (\%)},
        grid=major,
        axis background/.style={fill=scopePlotBg},
        grid style={draw=scopeGrid, dashed, line width=0.45pt},
        axis line style={draw=scopeAxis, line width=0.45pt},
        tick style={draw=scopeAxis, line width=0.45pt},
        tick align=outside,
        legend style={at={(0.5,1.035)},anchor=south,legend columns=3,draw=none,fill=none,inner xsep=1pt,inner ysep=1pt,font=\scriptsize,column sep=5pt,cells={anchor=west}},
        label style={font=\small},
        tick label style={font=\tiny},
        scaled ticks=false,
        clip=false
      ]
        \addplot+[scopeReact,line width=1.1pt,dashdotted,mark=*,mark size=1.15pt,mark options={solid,fill=scopeReact,draw=scopeReact}]
          coordinates {(0,12.22) (6,19.44) (12,21.11) (18,23.89) (24,25.00) (30,26.67)};
        \addlegendentry{ReAct}
        \addplot+[scopeReflexion,line width=1.1pt,dashed,mark=square*,mark size=1.15pt,mark options={solid,fill=scopeReflexion,draw=scopeReflexion}]
          coordinates {(0,15.56) (6,20.56) (12,24.44) (18,26.11) (24,27.78) (30,30.56)};
        \addlegendentry{Reflexion-Exp}
        \addplot+[scopeBlue,line width=1.35pt,mark=triangle*,mark size=1.35pt,mark options={solid,fill=scopeBlue,draw=scopeBlue}]
          coordinates {(0,16.11) (6,25.00) (12,33.33) (18,36.67) (24,39.44) (30,41.67)};
        \addlegendentry{SCOPE}
      \end{axis}
    \end{tikzpicture}
  \end{minipage}
  \hfill
  \begin{minipage}[t]{0.485\textwidth}
    \centering
    \begin{tikzpicture}
      \begin{axis}[
        width=0.98\linewidth,
        height=0.64\linewidth,
        xmin=0, xmax=30,
        ymin=10, ymax=44,
        xtick={0,6,12,18,24,30},
        ytick={10,15,20,25,30,35,40},
        xlabel={Number of simulation experiments},
        ylabel={ID Acc@1\% (\%)},
        grid=major,
        axis background/.style={fill=scopePlotBg},
        grid style={draw=scopeGrid, dashed, line width=0.45pt},
        axis line style={draw=scopeAxis, line width=0.45pt},
        tick style={draw=scopeAxis, line width=0.45pt},
        tick align=outside,
        legend style={at={(0.5,1.035)},anchor=south,legend columns=2,draw=none,fill=none,inner xsep=1pt,inner ysep=1pt,font=\scriptsize,column sep=8pt,cells={anchor=west}},
        label style={font=\small},
        tick label style={font=\tiny},
        scaled ticks=false,
        clip=false
      ]
        \addplot+[scopeBlue,line width=1.35pt,mark=triangle*,mark size=1.35pt,mark options={solid,fill=scopeBlue,draw=scopeBlue}]
          coordinates {(0,16.11) (6,25.00) (12,33.33) (18,36.67) (24,39.44) (30,41.67)};
        \addlegendentry{SCOPE}
        \addplot+[scopeSingle,line width=1.1pt,dashed,mark=diamond*,mark size=1.35pt,mark options={solid,fill=scopeSingle,draw=scopeSingle}]
          coordinates {(0,15.00) (6,22.78) (12,27.78) (18,30.56) (24,32.22) (30,33.89)};
        \addlegendentry{Single-Conjecture}
      \end{axis}
    \end{tikzpicture}
  \end{minipage}

    \caption{ID Acc@1\% with Qwen3.6-35B-A3B at 0, 6, 12, 18, 24, and 30 additional simulations. The six-call spacing follows SCOPE's two-round roundtables in these runs (three scientists, one simulation each per round). Left: ReAct, Reflexion-Exp, and SCOPE under the same simulation-call budget. Right: SCOPE and the budget-matched Single-Conjecture ablation.}
  \label{fig:acquisition_and_ablation}
  \label{fig:acquisition_dynamics}
  \label{fig:hypothesis_ablation_dynamics}
\end{figure*}

\subsection{Effect of Parallel Conjecture Exploration}
\label{sec:ablation}
The right panel of Figure~\ref{fig:hypothesis_ablation_dynamics} compares SCOPE with a budget-matched Single-Conjecture variant. Both methods use the same parallel branch count, initial observations, 30-call simulation budget, equation-search budget, and LLM-call limit. The variant follows a single primary conjecture, which is forcibly copied to other branches in each round, thereby eliminating conjecture diversity.

SCOPE and Single-Conjecture start from similar ID Acc@1\% levels, at 16.11\% and 15.00\%, respectively. As new evidence arrives, the gap widens from 1.11 percentage points initially to 7.78 percentage points at the final checkpoint, where SCOPE reaches 41.67\% and Single-Conjecture reaches 33.89\%. The widening gap may reflect how conjecture diversity shapes later investigation: branches starting from the same conjecture are more likely to pursue overlapping equation families and research questions, while distinct conjectures may expose different evidence gaps. Because SCOPE shares new observations across its branches, these varied questions can guide subsequent equation modeling and conjecture revision.

\FloatBarrier

\section{Conclusions and limitations}
This work highlights the value of developing scientific conjectures through sustained modeling and experimentation, with equations serving as both quantitative outcomes and tools for inquiry. SCOPE supports this process through parallel research trajectories and periodic evidence exchange. ClosureBench extends evaluation beyond equation recovery to the scientific findings and evidence produced during investigation. However, ClosureBench currently focuses on local algebraic closures in simulated PDE systems. Future work can extend this setting to differential, nonlocal, and history-dependent closures, which require richer models and experimental evidence.

Overall, SCOPE and ClosureBench suggest a direction for automated discovery: improving not only predictive models, but also how agents formulate questions, test explanations, and revise scientific understanding. Our work broadens the scope of conjecture-centered discovery and helps to assess how reliably autonomous systems can turn observations into scientific understanding.

\section*{The Use of Large Language Models (LLMs)}
LLMs serve several explicitly defined roles in this work. Within SCOPE, they act as the reasoning and decision-making components of the scientist agents, formulating and revising conjectures, guiding equation modeling, proposing research questions and experiments, interpreting evidence, and participating in roundtable discussions and voting, as described in Section~\ref{SCOPE}.

For ClosureBench construction, inspired by LLM-SRBench~\citep{llmsrbench}, we use LLMs to generate and structurally modify algebraic closure expressions embedded in known PDE systems. The resulting candidates undergo human review for dimensional consistency, physical constraints, and limiting behavior, together with numerical checks, as detailed in Section~\ref{sec:benchmark}.

For scientific reporting, an LLM synthesizes the task description, selected research trajectory, final equation, and accumulated evidence summary into a report. A reference report and writing guidelines provide organizational guidance, while scientific claims are required to be grounded in the current investigation, as described in Section~\ref{3.3}.

For evaluation, a fixed, method-blind LLM assesses the evidential support and scientific correctness of report claims for Evidence Faithfulness (EF) and Mechanism Insight (MI), respectively; the final scores are computed programmatically. Human validation involves two authors independently annotating 200 EF claims and 150 MI claim--fact pairs, with agreement reported in Appendix. Equation-accuracy metrics and Experimental Effectiveness (EE) do not rely on LLM judgments.

Beyond the use of LLMs in the method described above, LLMs served solely as auxiliary tools for generating some of the cartoon scientist characters in Figure~\ref{overview} and for polishing and refining the manuscript’s language. Their contributions to research conception, experimental design, or manuscript preparation do not warrant authorship or contributor status. All substantive research ideas, methods, and results presented in this work were independently developed or produced by the authors.

\bibliography{iclr2027_conference}
\bibliographystyle{iclr2027_conference}

\appendix

\section{Ethics and Privacy Statement}
ClosureBench is constructed from numerical simulations and does not contain personal, sensitive, or human-subject data.

\raggedbottom
\section{Sensitivity to Candidate-Set Capacities}
\label{app:meld-capacity-sensitivity}

To examine the sensitivity of SCOPE to candidate-set capacities, we fix the number of parallel scientists at \(M=3\) and vary the capacities of the candidate-generation and candidate-retention stages. In discovery round \(r\), for scientist \(i\), \(|\mathcal F_i^r|\) denotes the number of initial equation-family priors, \(|\widetilde{\mathcal F}_i^r|\) denotes the number of augmented equation candidates obtained after refinement, crossover and mutation, and \(K\) denotes the maximum number of elite equations retained for that scientist.

We use \((|\mathcal F_i^r|,|\widetilde{\mathcal F}_i^r|,K)=(4,12,4)\) as the default configuration and construct four variants around it. Prior-Narrow and Prior-Wide respectively decrease and increase candidate-generation capacity, whereas Retention-Narrow and Retention-Wide respectively decrease and increase the elite-equation capacity retained after validation. All configurations use the same 60 tasks, data and experimental budgets, stopping criteria, and three independent seeds.

We conduct this sensitivity analysis using Qwen as a representative backbone.

\begin{table}[H]
  \centering
  \caption{Sensitivity of SCOPE to candidate-set capacities with \(M=3\).}
  \label{tab:meld_capacity_sensitivity}
  \small
  \setlength{\tabcolsep}{6pt}
  \begin{tabular}{lcccccc}
    \toprule
    Setting
    & \(|\mathcal F_i^r|\)
    & \(|\widetilde{\mathcal F}_i^r|\)
    & \(K\)
    & ID Acc@1\% \(\uparrow\)
    & OOD Acc@1\% \(\uparrow\)
    & MI \(\uparrow\) \\
    \midrule
    Prior-Narrow     & 2 & 6  & 4 & 38.89 & 22.22 & 0.34 \\
    Default          & 4 & 12 & 4 & \textbf{41.67} & 25.00 & \textbf{0.38} \\
    Prior-Wide       & 8 & 24 & 4 & 40.56 & \textbf{25.56} & 0.36 \\
    Retention-Narrow & 4 & 12 & 2 & 39.44 & 21.67 & 0.31 \\
    Retention-Wide   & 4 & 12 & 8 & 41.11 & 23.89 & 0.35 \\
    \bottomrule
  \end{tabular}
\end{table}

For candidate generation, narrowing \((|\mathcal F_i^r|,|\widetilde{\mathcal F}_i^r|)\) reduces all three metrics, consistent with a search space that can discard useful equation families too early. Widening the generation space slightly improves OOD Acc, but ID Acc and MI both fall below the Default configuration. Thus, a larger candidate pool does not improve these three outcomes consistently.

For candidate retention, reducing \(K\) from 4 to 2 produces the clearest losses in OOD accuracy and MI. Because later investigation uses disagreements and residual structure among elite equations to formulate questions, retaining too few alternatives may limit discrimination among competing mechanisms. Increasing \(K\) to 8 also does not improve the results; additional retained equations may be redundant rather than consistently useful for subsequent modeling and experiment design.

Overall, the Default \((4,12,4)\) configuration provides the best balance across equation recovery, generalization, mechanism insight, and candidate-set size. Within the tested range, increasing capacity does not yield a consistent benefit.

\section{Representative Tasks in ClosureBench}
\label{app:representative_tasks}

To illustrate the construction of ClosureBench, we present one representative task from each of five scientific domains. These domains are reactive transport, energy and thermal sciences, environmental mass transfer, materials processing and rheology, and particulate and porous-media flow. In every task, a withheld closure relation is embedded in a host partial differential equation (PDE), and observations are generated by solving the resulting PDE system. A discovery method can access only the prescribed local variables and closure observations recovered from the discrete PDE residual.

The tasks use either standard constitutive forms or physics-informed benchmark variants. All reference closures are dimensionally consistent and satisfy the required conservation, positivity, and numerical-stability conditions over the declared operating ranges. Their numerical parameters are fixed as part of the benchmark specification to provide identifiable nonlinear targets and meaningful ID--OOD variation. Each closure contains at least one local input that varies over space and/or time within a trajectory; the remaining inputs may be spatial material fields or parameters fixed within one simulation.

Formally, the discovery target in each task is the complete symbolic mapping $g_\star$ from the observable closure inputs $\mathbf z$ to the closure output $y$, rather than the host PDE or a single scalar value:
\begin{equation*}
  y=g_\star(\mathbf z).
\end{equation*}
The method must recover both the functional structure and the numerical coefficients of this mapping from residual-recovered input--output pairs. The five mappings considered in this section are
\begin{align*}
  {S} &= {g_{\mathrm{reaction}}(C,T)}, \\
  {Q_J} &= {g_{\mathrm{Joule}}(T,E,\chi)}, \\
  {S_{\mathrm{abs}}} &= {g_{\mathrm{abs}}(C_g,C_b,d,T)}, \\
  {R_\psi} &= {g_{\mathrm{cryst}}(\psi,T,g)}, \\
  {g} &= {g_{\mathrm{exchange}}(C_m,C_i,r,\epsilon)}.
\end{align*}
In each task, the equation shown under ``Reference closure'' is the hidden ground-truth instance of the corresponding mapping and is used only by the simulator and evaluator.

\subsection{Task Overview and Common Evaluation Protocol}
\label{app:representative_tasks:overview}

\begin{table*}[t]
  \centering
  \caption{Representative ClosureBench tasks from five scientific domains. Experimental controls are simulator settings available to the investigating method; the closure inputs are the local variables used by the symbolic expression.}
  \label{tab:representative_tasks}
  \renewcommand{\arraystretch}{1.08}
  \setlength{\tabcolsep}{4pt}
  \begin{tabularx}{\textwidth}{
      >{\raggedright\arraybackslash}p{0.155\textwidth}
      >{\raggedright\arraybackslash}p{0.16\textwidth}
      >{\centering\arraybackslash}p{0.085\textwidth}
      >{\centering\arraybackslash}p{0.125\textwidth}
      >{\raggedright\arraybackslash}p{0.16\textwidth}
      >{\raggedright\arraybackslash}X}
    \toprule
    Scientific domain & Representative task & Closure output &
    Observable closure inputs & Role in the PDE & Experimental controls \\
    \midrule
    Reactive transport & One-sided heated reactive channel &
    $S$ & $C,T$ & Reaction source in the concentration balance &
    Inlet concentration, wall temperature, initial concentration \\
    \addlinespace[2pt]
    Energy and thermal sciences & Heterogeneous electrothermal heating &
    $Q_J$ & $T,E,\chi$ & Volumetric Joule source in the heat equation &
    Applied voltage, thermal level, composition-gradient amplitude \\
    \addlinespace[2pt]
    Environmental mass transfer & Counter-current spray-tower absorption &
    $S_{\mathrm{abs}}$ & $C_g,C_b,d,T$ & Coupled sinks in gas-solute and liquid-reagent balances &
    Inlet solute loading, nozzle diameter, inlet gas temperature \\
    \addlinespace[2pt]
    Materials processing and rheology & Flow-induced crystallization &
    $R_\psi$ & $\psi,T,g$ & Crystallinity source and latent-heat source &
    Inlet velocity, crystallinity seed, wall temperature \\
    \addlinespace[2pt]
    Particulate and porous-media flow & Mobile--immobile dual-porosity exchange &
    $g$ & $C_m,C_i,r,\epsilon$ & Opposite exchange source and sink in two pore regions &
    Darcy flux, inlet mobile concentration, initial immobile concentration, mean porosity \\
    \bottomrule
  \end{tabularx}
\end{table*}

\paragraph{Common evaluation protocol.} Each task specifies a public control box used to generate initial and additional discovery observations. ID test cases use held-out control settings within the same box. Every OOD test case has at least one control outside that box; other controls may remain within it. Fit, ID, and OOD observations are generated by complete OpenFOAM simulations. Only after an equation is locked are its predictions compared with closure observations reconstructed from the held-out discrete PDE balances.

\subsection{Local Reaction Source in a One-Sided Heated Channel}
\paragraph{Physical setting and discovery target.} A dilute reactant flows through a two-dimensional channel whose upper wall is heated while the lower wall remains at $300\,\mathrm{K}$. The velocity and pressure fields are solved to steady state and frozen while temperature $T$ and concentration $C$ evolve. The unknown is the complete source $S(C,T)$ in $\mathrm{mol\,m^{-3}\,s^{-1}}$. The public discovery/ID control box is $c_{\mathrm{in}}\in[0.5,1.2]\,\mathrm{mol\,m^{-3}}$, $T_{\mathrm{wall}}\in[310,340]\,\mathrm{K}$, and $C_0\in[0.05,0.50]\,\mathrm{mol\,m^{-3}}$.

\paragraph{Governing equation.} With velocity $\mathbf u$ and molecular diffusivity $D$, the concentration balance is
\[
\frac{\partial C}{\partial t}+\nabla\!\cdot(\mathbf u C)-\nabla\!\cdot(D\nabla C)=S(C,T).
\]

\paragraph{Reference closure.}
\[
S(C,T)=-0.025\frac{C}{C+0.5}\exp\!\left(\frac{T-300}{20}\right).
\]
The relation approaches first-order consumption at low concentration, saturates with concentration, and strengthens exponentially with temperature.

\paragraph{Observable variables and closure recovery.} Candidates use only beginning-of-step local $C$ and $T$. The observed source is recovered from the complete discrete concentration balance after subtracting convection and diffusion.

\paragraph{OOD test ranges.} The low-side out-of-range control values are $c_{\mathrm{in}}\in[0.395,0.5)\,\mathrm{mol\,m^{-3}}$, $T_{\mathrm{wall}}\in[305.5,310)\,\mathrm{K}$, and $C_0\in[0.014,0.05)\,\mathrm{mol\,m^{-3}}$; the high-side values are $c_{\mathrm{in}}\in(1.2,1.305]\,\mathrm{mol\,m^{-3}}$ and $T_{\mathrm{wall}}\in(340,344.5]\,\mathrm{K}$.

\subsection{Heterogeneous Electrothermal Heating}
\paragraph{Physical setting and discovery target.} A voltage drives current through a heterogeneous two-dimensional conducting sheet while its boundary is held at a prescribed thermal level. The local composition field $\chi$ varies across the sheet. The target is the positive volumetric Joule source $Q_J(T,E,\chi)$ in $\mathrm{W\,m^{-3}}$, where $T$ is temperature in kelvin and $E=|\nabla\phi|$ is electric-field magnitude in $\mathrm{V\,m^{-1}}$. The public control box is applied voltage $V\in[4,9]\,\mathrm{V}$, thermal level $T_b\in[295,325]\,\mathrm{K}$, and composition-gradient amplitude $\gamma_\chi\in[-0.36,0.36]$.

\paragraph{Governing equations.} The conductivity $\sigma(T,\chi)$ determines both the electric-potential solve and the heat source:
\[
\nabla\!\cdot\!\left(\sigma(T,\chi)\nabla\phi\right)=0,
\qquad E=|\nabla\phi|,
\]
\[
\rho C_p\frac{\partial T}{\partial t}-\nabla\!\cdot(k\nabla T)=Q_J(T,E,\chi).
\]

\paragraph{Reference closure.}
\[
Q_J(T,E,\chi)
=\left(1\,\mathrm{S\,m^{-1}}\right)
 \left(\frac{T}{300\,\mathrm{K}}\right)^2(0.1+4\chi)E^2.
\]
Thus $Q_J=\sigma E^2$ with the same positive conductivity used in the potential equation.

\paragraph{Observable variables and closure recovery.} Candidates use only the step-start local $T$, $E$, and $\chi$. The observed $Q_J$ is reconstructed from the complete discrete transient-conduction balance after the potential equation has supplied $E$.

\paragraph{OOD test ranges.} The low-side out-of-range control ranges are $V\in[2.8,3.0]\,\mathrm{V}$, $T_b\in[288,289]\,\mathrm{K}$, and $\gamma_\chi\in[-0.4896,-0.4464]$; the high-side ranges are $V\in[10.0,10.2]\,\mathrm{V}$, $T_b\in[331,332]\,\mathrm{K}$, and $\gamma_\chi\in[0.4464,0.4896]$.

\subsection{Counter-Current Spray-Tower Absorption}
\paragraph{Physical setting and discovery target.} Acid-bearing gas rises through falling alkaline droplets. The target $S_{\mathrm{abs}}(C_g,C_b,d,T)$ is the positive absorption rate in $\mathrm{s^{-1}}$. Here $C_g$ and $C_b$ are normalized gas-solute and liquid-reagent concentrations, $d=d_{\mathrm{local}}/(0.0016\,\mathrm{m})$, and $T=T_{\mathrm{kelvin}}/(323.15\,\mathrm{K})$. The public control box is inlet solute loading $L\in[0.45,1.55]$, nozzle diameter $d_n\in[0.0009,0.0026]\,\mathrm{m}$, and inlet gas temperature $T_{g,\mathrm{in}}\in[323,353]\,\mathrm{K}$.

\paragraph{Governing equations.} With gas and liquid velocities $\mathbf U_g$ and $\mathbf U_b$, dispersivities $D_g$ and $D_b$, and known reagent coupling $\kappa$,
\[
\frac{\partial C_g}{\partial t}+\nabla\!\cdot(\mathbf U_g C_g)-\nabla\!\cdot(D_g\nabla C_g)=-S_{\mathrm{abs}},
\]
\[
\frac{\partial C_b}{\partial t}+\nabla\!\cdot(\mathbf U_b C_b)-\nabla\!\cdot(D_b\nabla C_b)=-\kappa S_{\mathrm{abs}}.
\]

\paragraph{Reference closure.}
\[
S_{\mathrm{abs}}
=0.3\,\mathrm{s^{-1}}
\frac{C_gC_b}{C_b+0.35T^{12}}\,d^{-1.7}\exp[-4(T-1)].
\]
The rate vanishes with either reactant, decreases for coarser droplets, and combines reagent saturation with a temperature-dependent absorption factor.

\paragraph{Observable variables and closure recovery.} The gas-solute and liquid-reagent balances independently reconstruct $S_{C_g}$ and $S_{C_b}$. The published label is their mean, $S_{\mathrm{observed}}=(S_{C_g}+S_{C_b})/2$, in $\mathrm{s^{-1}}$. The inputs are normalized as defined above, while the target retains the physical multiplier $S_{\mathrm{ref}}=0.3\,\mathrm{s^{-1}}$.

\paragraph{OOD test ranges.} The low-side out-of-range control ranges are $L\in[0.27,0.30]$, $d_n\in[0.00063,0.00066]\,\mathrm{m}$, and $T_{g,\mathrm{in}}\in[318,319]\,\mathrm{K}$; the high-side ranges are $L\in[1.70,1.73]$, $d_n\in[0.00284,0.00287]\,\mathrm{m}$, and $T_{g,\mathrm{in}}\in[357,358]\,\mathrm{K}$.

\subsection{Flow-Induced Crystallization}
\paragraph{Physical setting and discovery target.} A polymer melt flows through a cooled channel. A frozen velocity field supplies the local shear rate while crystallinity $\psi\in[0,1]$ and absolute temperature $T$ evolve. The target is the nonnegative crystallization rate $R_\psi(\psi,T,g)$ in $\mathrm{s^{-1}}$, with normalized shear $g=\dot\gamma/(10\,\mathrm{s^{-1}})$. The public control box is inlet velocity $u_{\mathrm{in}}\in[0.012,0.048]\,\mathrm{m\,s^{-1}}$, crystallinity seed $\psi_0\in[0.06,0.35]$, and wall temperature $T_w\in[330,390]\,\mathrm{K}$.

\paragraph{Governing equations.} With crystallinity diffusivity $D_\psi$, thermal diffusivity $\alpha$, and the known latent-temperature coefficient $18\,\mathrm{K}$,
\[
\frac{\partial\psi}{\partial t}+\nabla\!\cdot(\mathbf U\psi)-\nabla\!\cdot(D_\psi\nabla\psi)=R_\psi,
\]
\[
\frac{\partial T}{\partial t}+\nabla\!\cdot(\mathbf U T)-\nabla\!\cdot(\alpha\nabla T)=18\,\mathrm{K}\,R_\psi.
\]

\paragraph{Reference closure.}
\[
R_\psi(\psi,T,g)
=0.055\,\mathrm{s^{-1}}\,27(1-\psi)^{1.5}
 \log\!\left(\frac{T}{320\,\mathrm{K}}\right)
 \log\!\left(\frac{440\,\mathrm{K}}{T}\right)(1+g).
\]
The two logarithms create an endpoint-zero temperature window between the glass-transition and melting temperatures, while shear promotes crystallization.

\paragraph{Observable variables and closure recovery.} Candidates use step-start $\psi$, temperature $T$ in kelvin, and normalized shear $g$. The crystallinity balance directly reconstructs $R_\psi$, while the temperature balance divided by $18\,\mathrm{K}$ provides an independent check. The published label retains the physical factor $R_{\mathrm{ref}}=0.055\,\mathrm{s^{-1}}$.

\paragraph{OOD test ranges.} The low-side out-of-range control ranges are $u_{\mathrm{in}}\in[0.0063,0.0069]\,\mathrm{m\,s^{-1}}$, $\psi_0\in[0.028,0.034]$, and $T_w\in[322,324]\,\mathrm{K}$; the high-side ranges are $u_{\mathrm{in}}\in[0.0531,0.0537]\,\mathrm{m\,s^{-1}}$, $\psi_0\in[0.376,0.382]$, and $T_w\in[396,398]\,\mathrm{K}$.

\subsection{Mobile--Immobile Exchange in a Dual-Porosity Medium}
\paragraph{Physical setting and discovery target.} A tracer moves through connected mobile pores while adjacent dead-end pores hold an immobile inventory. The discovery target is the dimensionless normalized exchange closure $g(C_m,C_i,r,\epsilon)=S_{\mathrm{physical}}/S_{\mathrm{ref}}$, where $C_m$ and $C_i$ are normalized local concentrations, $r=|\mathbf q|/[\epsilon(0.001\,\mathrm{m\,s^{-1}})]$, $\epsilon$ is mobile porosity, and $S_{\mathrm{ref}}=0.002\,\mathrm{s^{-1}}$. The public control box is Darcy flux $q\in[0.0002,0.0008]\,\mathrm{m\,s^{-1}}$, inlet mobile concentration $C_{m,\mathrm{in}}\in[0.2,1.6]$, initial immobile concentration $C_{i,0}\in[0.1,1.5]$, and mean porosity $\bar\epsilon\in[0.36,0.46]$.

\paragraph{Governing equations.} With mobile and immobile diffusivities $D_m$ and $D_i$, the physical source $S_{\mathrm{physical}}=S_{\mathrm{ref}}g$ enters the two balances with opposite signs:
\[
\frac{\partial(\epsilon C_m)}{\partial t}+\nabla\!\cdot(\mathbf q C_m)-\nabla\!\cdot(\epsilon D_m\nabla C_m)=-S_{\mathrm{physical}},
\]
\[
\frac{\partial[(1-\epsilon)C_i]}{\partial t}-\nabla\!\cdot[(1-\epsilon)D_i\nabla C_i]=+S_{\mathrm{physical}}.
\]

\paragraph{Reference closure.}
\[
g(C_m,C_i,r,\epsilon)
=(C_m-C_i)\epsilon(1-\epsilon)
\left(\frac{1}{1+r}+\frac{1}{1+C_m+C_i}\right).
\]
Positive values transfer solute from mobile to immobile pores; negative values remobilize it.

\paragraph{Observable variables and closure recovery.} The mobile and immobile balances independently reconstruct the same signed physical exchange. The observed target is the mean recovered source divided by $S_{\mathrm{ref}}$, so candidate expressions recover the dimensionless $g$, not the physical source $S_{\mathrm{physical}}$.

\paragraph{OOD test ranges.} The low-side out-of-range control ranges are $q\in[0.00012,0.00014]\,\mathrm{m\,s^{-1}}$ and $C_{m,\mathrm{in}}\in[0.05,0.08]$; the high-side ranges are $q\in[0.00086,0.00089]\,\mathrm{m\,s^{-1}}$ and $C_{m,\mathrm{in}}\in[1.72,1.76]$. Across these OOD cases, $C_{i,0}\in[0.24,1.36]$ and $\bar\epsilon\in[0.37,0.45]$ remain within their public ranges.

\section{Implementation Details}
\label{app:implementation-details}

\subsection{Computational Resources and Software}
\label{app:compute-environment}

We conduct the experiments on a computing cluster equipped with 8 NVIDIA H800 GPUs and 8 NVIDIA H200 GPUs. The PDE
solvers and interactive computational experiments in ClosureBench are implemented with OpenFOAM v2512. PySR and all
method-specific invocations of it use PySR 2.0.0a2. Unless otherwise specified below, we retain the default settings of the
corresponding public implementations.

\subsection{Baseline Implementations and Hyperparameters}
\label{app:baseline-configurations}
Symbolic regression (SR) has received growing attention for its ability to discover interpretable mathematical expressions from data~\citep{chen2022symbolic,xu2023retention,meidani2023snip,li2023discrete,e2e,li2024focus,li2026learning,hu2026dass,li2026score}. Beyond numerical prediction, SR turns observations into testable scientific hypotheses, supporting equation discovery across physics, biology, materials science, and other domains~\citep{yu2025discover,li2023adaptive,hu2024domain,guo2025sr,hu2025context,xu2025exploring,xu2024worth,xiong2026enhancing,xiong2025deliberate}. Among the diverse SR paradigms, we select representative methods from different solver families as baselines.

\paragraph{End-to-End Symbolic Regression (E2E)}
We follow the public E2E implementation~\cite{e2e}, providing the pretrained Transformer with at most 200 observations and
using 10 candidate trees for refinement on each task.

\paragraph{Deep Symbolic Regression (DSR)}
We follow the public DSR implementation~\cite{dsr}, which searches for expressions using risk-seeking policy gradients. For
each task, we sample $100{,}000$ candidates with a batch size of 512 and a learning rate of $5\times10^{-4}$. The operator
set contains addition, subtraction, multiplication, division, sine, cosine, exponential, and logarithmic functions.

\paragraph{Unified Deep Symbolic Regression (uDSR)}
We augment the DSR configuration with the genetic-programming component of uDSR. We use a population of 100 for 20
generations, set both the crossover and mutation probabilities to 0.5, use a tournament size of 5, and limit mutated subtrees
to a maximum depth of 3. All remaining optimization parameters follow those of DSR.

\paragraph{Python Symbolic Regression (PySR)}
We configure PySR~\cite{pysr} with 15 populations of 33 individuals and run it for 125 iterations, each containing 550
evolutionary cycles. We set the maximum expression size to 50 and the randomization weight to 0.1, and select the output
expression using \texttt{model\_selection=best}. The binary operator set is
$\{+,-,\times,\div,\operatorname{pow}\}$, and the unary operator set is
$\{\exp,\log,\sqrt{\cdot},\sin,\cos\}$. Search is executed with multithreading. To exclude pathological compositions, we
disallow mutual nesting between trigonometric functions, mutual nesting between exponential and logarithmic functions, and
self-nesting of square roots, while imposing subtree-complexity bounds on the nonlinear operators.

\paragraph{Library-Augmented Symbolic Regression (LaSR)}
We follow the LaSR configuration used in SR-Scientist~\cite{xia2025sr}. We configure LaSR~\cite{lasr} with 10 populations of 33 individuals and run it for 25 iterations, each containing 550
evolutionary cycles. The maximum expression size is 30, and the concept library retains at most 20 concepts. The weights of
the LLM-based crossover, mutation, and randomization operations are each set to 0.02. Its symbolic operators and nesting
constraints are identical to those used by PySR.

\paragraph{LLM-SR}
We follow the original multi-island search procedure of LLM-SR~\cite{llmsr}. For each task, we use 10 islands and generate 1,000 candidate samples. We adopt the NumPy+BFGS configuration. Each expression contains at most 10 free parameters, optimized to minimize the mean squared error on the observed data. We retain the original prompts, experience buffer, intra-island search procedure, and candidate ranking.

\paragraph{SR-Scientist}
We use the public SR-Scientist framework~\cite{xia2025sr}, which performs long-horizon candidate optimization with a data
analyzer and an equation evaluator. We allow at most 25 tool-interaction turns per outer iteration and at most 40 outer
iterations. The initial MAPE target is $0.1\%$, and the termination threshold is $0.0001\%$. The equation evaluator optimizes
at most 10 free parameters using BFGS with mean squared error as its objective.

\paragraph{ReAct}
Following ReAct~\cite{yao2022react}, we maintain a single reasoning--action trajectory. At each step, the model selects its
next action based on the current data, existing candidates, and previous tool observations, and continues reasoning after
receiving the new observation. Each action step selects exactly one tool and is not equivalent to one simulation experiment. Available actions include data analysis, literature retrieval, PySR, LLM-SR, candidate
execution and evaluation, and the submission of admissible OpenFOAM experiments. The agent workflow requires an SR-tool call before the first additional simulation. Our implementation introduces no additional planner, reflection module, or
cross-round scientific memory.

\paragraph{Reflexion-Exp}
Following Reflexion~\cite{shinn2023reflexion}, we adapt its reflection-memory mechanism to multi-round computational
experimentation. Within each round, Reflexion-Exp retains the same reasoning--action process and tool interfaces as ReAct. Reflexion-Exp follows the same initial SR-tool requirement as ReAct.
After every three successful simulation experiments, the model generates a reflection from the new observations, changes in the
candidate set, and recent feedback, and appends it to persistent cross-round memory. The reflection call counts toward the method's LLM budget. Subsequent action steps condition on both
the current evidence and the stored reflections, allowing the agent to revise earlier strategies and avoid repeated failures.

\subsection{Shared Literature-Retrieval Tool}
\label{app:literature-retrieval}

All interactive agent methods are equipped with the same cross-disciplinary scholarly search interface and the same maximum retrieval budget. For each retrieval call, the agent generates structured queries from the task description, the currently available public evidence, and its method-specific discovery state, and retrieves paper metadata through the Semantic Scholar Academic Graph~\cite{kinney2023semantic}. Following established scientific literature retrieval and attribution pipelines~\cite{singh-etal-2025-ai2,chen-etal-2025-surveygen}, the system deduplicates and ranks the retrieved papers by relevance and primarily processes their titles and abstracts. When verified open-access full text is available, the system additionally extracts relevant passages; otherwise, it retains abstract-level evidence. The retrieved evidence, together with its source information, is then added to the method's public evidence context.

\section{Detailed Algorithm of SCOPE}
\label{app:meld-algorithm}

Table~\ref{alg:meld} specifies the conjecture-centered loop used by SCOPE. SCOPE uses $M=3$ scientists, $T=10$ discovery rounds, and a roundtable after every two rounds. Each scientist maintains its own conjecture, candidate-equation collection, question record, and research log, while all scientists use the shared observation set.

\begin{longtblr}[
  caption={The conjecture-centered scientific discovery procedure of SCOPE.},
  label={alg:meld}
]{colspec={Q[r,wd=3.5em] X[l]},colsep=3pt,rowsep=0pt}
    \toprule
    \textbf{Input:} &
      Initial observations $\mathcal D_0$; scientific context $\mathcal C$; rounds $T=10$; scientists $M=3$; SR solvers $\mathcal S$; admissible experiments $\mathcal A$.\\
    \textbf{Output:} &
      Selected equation $f^\star$; evidence-grounded scientific report $\mathcal R$.\\
    \midrule
    1 & For each scientist $i$, analyze $\mathcal D_0$, retrieve relevant literature, initialize $H_i^0$, and set $\mathcal V_i^0=\mathcal Q_i^0=\varnothing$ with the initial analyses recorded in $\mathcal L_i^0$.\\
    2 & \textbf{for} $r=1,\ldots,T$ \textbf{do}\\
    3 & \quad \textbf{for each} scientist $i=1,\ldots,M$ \textbf{in parallel do}\\
    4 & \qquad Expand the equation families in $H_i^{r-1}$ and run the solvers in $\mathcal S$ on $\mathcal D_{r-1}$, seeded by accessible candidates in $\mathcal V_i^{r-1}$.\\
    5 & \qquad Filter invalid or duplicate equations, evaluate all retained candidates, select the elite set $\mathcal E_i^r$, and update $\mathcal V_i^r$.\\
    6 & \qquad Use $H_i^{r-1}$, $\mathcal E_i^r$, $\mathcal Q_i^{r-1}$, and $\mathcal L_i^{r-1}$ to select an unresolved question from prediction disagreements, residual structure, or unsupported ranges.\\
    7 & \qquad Run targeted analyses; propose an admissible experiment with required measurements, expected outcomes, and falsification criteria.\\
    8 & \quad \textbf{end for}\\
    9 & \quad Execute the feasible proposed experiments within the shared remaining budget, obtaining $\mathcal A_r$ and observations $\mathcal O(a)$.\\
    10 & \quad Update the shared data:
      $\mathcal D_r\leftarrow\mathcal D_{r-1}\cup\bigcup_{a\in\mathcal A_r}\mathcal O(a)$.\\
    11 & \quad \textbf{for each} scientist $i=1,\ldots,M$ \textbf{in parallel do}\\
    12 & \qquad Compare the returned evidence with the registered expected outcomes and falsification criteria; record the result as support, refutation, or inconclusive.\\
    13 & \qquad Revise $H_i^r$, update the statuses and evidence links in $\mathcal Q_i^r$, and append the analyses, experiment assessment, and revision to $\mathcal L_i^r$.\\
    14 & \quad \textbf{end for}\\
    15 & \quad Re-evaluate each $\mathcal E_i^r$ on the updated observations $\mathcal D_r$.\\
    16 & \quad \textbf{if} $r$ is even \textbf{then}\\
    17 & \qquad Hold a roundtable in which all scientists share $(H_i^r,\mathcal E_i^r,\mathcal Q_i^r,\mathcal L_i^r)$ and independently vote for the most convincing trajectory.\\
    18 & \qquad Record every vote and rationale. If all votes select the same trajectory $w$, set the stopping round $R\leftarrow r$ and \textbf{break}.\\
    19 & \qquad If rounds remain, let each scientist add a question prompted by another trajectory's conjecture, models, or evidence.\\
    20 & \quad \textbf{end if}\\
    21 & \textbf{end for}\\
    22 & If no unanimous winner was found, set $R\leftarrow T$ and choose $w$ by plurality in the fifth roundtable; break ties by the lower mean prediction error of the tied trajectories.\\
    23 & Select $f^\star\leftarrow\arg\min_{f\in\mathcal E_w^R}\ell(f;\mathcal D_R)$ from the winning trajectory's final elite set.\\
    24 & Generate $\mathcal R$ from the selected conjecture and equation, their supporting or refuting evidence, scope and unresolved questions, and the retained alternative trajectories.\\
    25 & \textbf{return} $(f^\star,\mathcal R)$.\\
    \bottomrule
\end{longtblr}

\definecolor{scopePromptFill}{rgb}{0.95,0.97,1.0}
\definecolor{scopePromptLine}{rgb}{0.2,0.4,0.7}
\newcommand{\scopePromptRole}{Prompt}
\newcommand{\role}[1]{\def\scopePromptRole{#1}}
\tcbset{scope prompt/.style={
  enhanced,breakable,
  colback=scopePromptFill,colframe=scopePromptLine,
  colbacktitle=scopePromptFill,coltitle=black,coltext=black,
  boxrule=.6pt,arc=2mm,
  left=2.5mm,right=2.5mm,top=1.5mm,bottom=1.5mm,
  toptitle=1.2mm,bottomtitle=1.2mm,
  fonttitle=\small\bfseries,
  before skip=7pt,after skip=8pt,
  pad at break*=1.5mm,
  title={\scopePromptRole},
  title after break={\scopePromptRole\ (continued)}
}}
\newenvironment{promptblock}{%
  \VerbatimEnvironment
  \begin{tcolorbox}[scope prompt]
  \begin{Verbatim}[fontsize=\fontsize{8}{9.6}\selectfont,
    breaklines=true,breakanywhere=true,
    breaksymbolleft={},breaksymbolright={},
    breakanywheresymbolpre={},breakanywheresymbolpost={},
    formatcom=\color{black},baselinestretch=1]%
}{%
  \end{Verbatim}
  \end{tcolorbox}
}
\newtcolorbox{programrule}{scope prompt,title=Program-side rules,
  title after break={Program-side rules (continued)},fontupper=\small}

\section{SCOPE Prompts}
\label{app:scope-prompts}

This appendix presents the main prompts used across SCOPE's research process: structured conjecture initialization; question selection, analysis, and experiment proposal; evidence assessment and conjecture revision; roundtable exchange and independent voting; and scientific report generation. A conjecture corresponds to \(H_i^r\) in the method section, comprising a mechanistic explanation, relevant variables, qualitative relationships, testable predictions, candidate equation families, and an evidence summary. Candidate expressions are quantitative realizations developed and tested under a conjecture. Angle-bracketed text denotes information supplied at run time.

\subsection{Structured conjecture initialization}
Initialization has two levels. A shared SCOPE call generates candidate equations and structured conjecture drafts from public starting evidence. Each scientist then independently confirms and refines one assigned draft using the same evidence, establishing its initial research direction.

\subsubsection{Shared candidates and conjecture drafts}
\textbf{Call purpose.} Produce the shared candidate foundation and exactly three genuinely distinct structured conjecture drafts from public evidence.

\role{System message}
\begin{promptblock}
You are the SCOPE initialization component.

Construct a shared foundation of executable closure candidates and three structured scientific conjecture drafts from public evidence. A conjecture is a scientific explanation that guides equation modeling, prediction, and investigation; it is not merely an equation or an equation cluster.

The unknown closure may be recovered from public evidence. Never access or use hidden Truth, private ID/OOD evaluation, or any private benchmark evidence. Do not present a well-fitting expression as a confirmed mechanism without supporting public evidence. Return concise scientific rationale and evidence references, not chain-of-thought.
\end{promptblock}

\role{User message}
\begin{promptblock}
PUBLIC TASK CARD
<PUBLIC_TASK_CARD>

PUBLIC INITIAL OBSERVATIONS AND SUMMARIES
<PUBLIC_INITIAL_EVIDENCE_JSON>

OPTIONAL PUBLIC LITERATURE EVIDENCE
<OPTIONAL_PUBLIC_LITERATURE_JSON>

EXISTING INITIAL CANDIDATES
<INITIAL_CANDIDATES_JSON>

Execute the shared candidate-generation and structural-expansion stages: scientific priors, mechanism grouping into exactly three conjecture drafts, parent equation families, scientifically meaningful mutation, and compatible recombination. Preserve useful existing candidates. Assign an existing candidate to every compatible conjecture_id; if it is not specific to one conjecture, keep it global/shared.

Return exactly three conjecture drafts with genuinely different mechanistic stories and functional implications, not coefficient variants or cosmetic transformations of one family. Each draft must contain the complete H structure:
- conjecture_id
- name
- mechanistic_claim
- relevant_variables
- qualitative_relations
- testable_prediction
- family_expressions
- compatible_initial_candidate_ids
- status
- evidence_summary

For every candidate, return:
- name, expression, conjecture_id
- operator, parent_candidate_ids, lineage
- mechanism, expected_trends, limiting_behavior, scientific_risk

Use only public input fields and operators allowed by the task card. Keep equations complete and executable, with numeric constants where required. Make the predictions specific enough to be checked against observations. In evidence_summary, distinguish what public evidence supports, what challenges the draft, and what remains untested.
\end{promptblock}

\textbf{Output format.} One JSON object contains scientific priors, exactly three conjecture records, and a candidate-equation array. Each conjecture record includes the six scientific components of the structured conjecture \(H\) defined in the main text.

\subsubsection{Independent scientist confirmation and refinement}
\textbf{Call purpose.} Let each scientist independently accept, revise, or sharpen its assigned draft before its trajectory begins.

\role{System message}
\begin{promptblock}
You are one scientist in SCOPE. Independently confirm and refine one assigned structured conjecture using public evidence. Preserve the complete H structure and the candidate foundation. Do not revise another scientist's conjecture.
\end{promptblock}

\role{User message}
\begin{promptblock}
SCIENTIST
<SCIENTIST_ID>

PUBLIC TASK CARD AND INITIAL EVIDENCE
<PUBLIC_TASK_CARD>
<PUBLIC_INITIAL_EVIDENCE_JSON>
<OPTIONAL_PUBLIC_LITERATURE_JSON>

YOUR ASSIGNED CONJECTURE DRAFT
<ASSIGNED_CONJECTURE_JSON>

SHARED CANDIDATE-GENERATION AND STRUCTURAL-EXPANSION FOUNDATION
<SHARED_CANDIDATES_JSON>

Independently examine whether the assigned mechanistic claim, variable set, qualitative relations, predictions, and equation families are mutually consistent and grounded in the public evidence. Refine weak or overbroad parts. Keep conjecture_id unless the draft is internally incoherent; if it must change, explain the scientific reason briefly.

Return one complete H object using the established fields, plus:
- scientific_rationale: a short justification for the confirmed or revised direction
- evidence_refs: public evidence, candidate, or literature identifiers actually used

Do not report private reasoning traces. Do not claim confirmation where the evidence only supports a testable possibility.
\end{promptblock}

\textbf{Output contract.} \texttt{conjectures} contains exactly one complete record; \texttt{scientific\_rationale} and \texttt{evidence\_refs} accompany it.

\subsection{Question-driven analysis and experiment proposal}
After selecting a question, the scientist first analyzes existing evidence, then proposes an experiment to address an important question that remains unresolved.

\subsubsection{Select the unresolved scientific question}
\role{System message}
\begin{promptblock}
Select one consequential, investigable scientific question from public evidence.
\end{promptblock}

\role{User message}
\begin{promptblock}
You are scientist <SCIENTIST_ID> in SCOPE round <ROUND_INDEX>.

PUBLIC TASK AND SHARED OBSERVATIONS
<PUBLIC_TASK_CARD>
<PUBLIC_OBSERVATIONS_JSON>

YOUR COMPLETE CURRENT H
<CURRENT_CONJECTURE_JSON>

YOUR ELITE EQUATIONS AND PUBLIC EVALUATIONS
<ELITE_CANDIDATES_JSON>

YOUR QUESTION AND RESEARCH RECORDS
<QUESTION_HISTORY_JSON>
<RESEARCH_RECORD_JSON>

QUESTION DIAGNOSTICS
<CANDIDATE_DISAGREEMENTS_JSON>
<RECURRING_RESIDUAL_PATTERNS_JSON>
<UNSUPPORTED_PREDICTIONS_OR_RANGES_JSON>

Select an existing open question or propose one new question whose resolution could change, constrain, or strengthen your conjecture. Ground it in candidate disagreement, residual structure, unsupported ranges or predictions, parameter instability, identifiability, or a competing conjecture prediction.

Return:
- questions: up to <MAX_CANDIDATE_QUESTIONS>, each with text, basis,
basis_evidence_refs, and source_type
- selected_index and why_now
- candidate_evidence_comparison
- evidence_gap
- selection_reason
- evidence_refs

State what is not yet known. Do not turn the question into a conclusion and do not yet request an experiment.
\end{promptblock}

\subsubsection{Analyze existing evidence before proposing an experiment}
\role{System message}
\begin{promptblock}
Investigate the selected question with existing public evidence and approved analysis tools. Determine what is already resolved and what evidence gap remains.
\end{promptblock}

\role{User message}
\begin{promptblock}
SELECTED QUESTION
<SELECTED_QUESTION_JSON>

CURRENT H, ELITE EQUATIONS, AND PUBLIC OBSERVATIONS
<CURRENT_CONJECTURE_JSON>
<ELITE_CANDIDATES_JSON>
<PUBLIC_OBSERVATIONS_JSON>

AVAILABLE ANALYSIS RESULTS
<DATA_ANALYZER_RESULTS_JSON>
<EXISTING_ROW_CONTRAST_RESULTS_JSON>
<EQUATION_PROPERTY_CHECK_RESULTS_JSON>

Use the available results to address the selected question. Compare residuals and input coverage, within- and across-condition contrasts, and relevant equation properties such as dependencies, interactions, local behavior, parameter stability, and identifiability.

Return:
- analysis_conclusion
- resolved_parts
- remaining_gap
- candidate_implications
- required_observation
- scientific_rationale
- evidence_refs

If existing evidence already resolves the question, say so and set required_observation to null. Otherwise describe the observation needed, without pretending that directly selecting a local state is the same as choosing executable experimental conditions.
\end{promptblock}

\subsubsection{Propose an executable, discriminating experiment}
\role{System message}
\begin{promptblock}
Translate a remaining scientific evidence gap into an executable experiment proposal. Register predictions before the experiment runs.
\end{promptblock}

\role{User message}
\begin{promptblock}
PUBLIC TASK AND FEASIBLE CONTROL INTERFACE
<PUBLIC_TASK_CARD>
<FEASIBLE_CONTROL_SPACE_JSON>

SELECTED QUESTION AND COMPLETED ANALYSIS
<SELECTED_QUESTION_JSON>
<ANALYSIS_RECORD_JSON>

CURRENT H AND ELITE EQUATIONS
<CURRENT_CONJECTURE_JSON>
<ELITE_CANDIDATES_JSON>

If the completed analysis leaves a consequential gap, propose one feasible experiment that targets it. Specify controllable conditions rather than unattainable local states. Register what must be measured, what each plausible outcome would mean, and what would challenge the current conjecture. If no new experiment is warranted, return experiment_required=false and identify the existing evidence that resolves the question.

Return:
- question_id
- experiment_required
- experiment_conditions
- required_measurements
- registered_predictions, each with prediction_id, prediction, applicable_window,
expected_outcome, and interpretation
- falsification_criteria
- feasibility_rationale
- selection_reason
- evidence_refs
\end{promptblock}

\subsection{Program verdicts and complete conjecture revision}
The model does not invent a fresh post hoc verdict. It reads the program-computed assessment of each registered prediction under its specified window and criterion, then revises the complete structured conjecture from the combined evidence.

\role{System message}
\begin{promptblock}
Revise one complete structured conjecture from public evidence and program-computed assessments of registered predictions. Preserve evidence limits and unresolved questions.
\end{promptblock}

\role{User message}
\begin{promptblock}
You are scientist <SCIENTIST_ID> after SCOPE round <ROUND_INDEX>.

YOUR COMPLETE PRE-EXPERIMENT H
<CURRENT_CONJECTURE_JSON>

ELITE EQUATIONS AND PRIOR ANALYSIS
<ELITE_CANDIDATES_JSON>
<ANALYSIS_RECORD_JSON>

EXPERIMENT REQUEST AND RETURNED PUBLIC OBSERVATIONS
<REGISTERED_EXPERIMENT_JSON>
<RETURNED_PUBLIC_OBSERVATIONS_JSON>

PROGRAM-COMPUTED VERDICTS
<PROGRAM_COMPUTED_VERDICTS_JSON>

Read the computed verdict for each registered prediction. Treat supported as evidence for that prediction in its specified window, refuted as a challenge to that prediction under its specified criterion, and inconclusive as unresolved. These verdicts concern registered predictions, not the whole conjecture; revision of the complete conjecture follows from the combined evidence. A not_obtained result is inconclusive: the required observation was not reached, so it is not a refutation.

Update the complete H object: conjecture_id, name, mechanistic_claim, relevant_variables, qualitative_relations, testable_prediction, family_expressions, compatible_initial_candidate_ids, status, and evidence_summary. Revise only the parts affected by evidence. Preserve viable alternatives and explicit limits. Keep conjecture_id unless the mechanism itself changes.

Also return:
- assessment: prediction_id, observed_result, verdict, and evidence_refs
- revision_summary: changed_parts, retained_parts, and unresolved_parts
- scientific_rationale
- next_open_questions
- evidence_refs

Do not recompute or overrule the verdicts. Do not expose chain-of-thought.
\end{promptblock}

\subsection{Roundtable exchange and independent vote}
Every two rounds, each scientist reads the same snapshot of all trajectories, including its own. Exchange may influence later questions and candidate use, but the prompt does not force a conjecture replacement or perform the program's vote count and final selection.

\role{System message}
\begin{promptblock}
Compare all SCOPE trajectories from one shared snapshot. Independently vote for the most convincing trajectory and identify what should inform your next investigation.
\end{promptblock}

\role{User message}
\begin{promptblock}
You are scientist <SCIENTIST_ID> at roundtable <ROUNDTABLE_INDEX>.

PUBLIC TASK AND CUMULATIVE PUBLIC OBSERVATIONS
<PUBLIC_TASK_CARD>
<CUMULATIVE_PUBLIC_OBSERVATIONS_JSON>

SAME ROUNDTABLE SNAPSHOT FOR EVERY SCIENTIST
<ALL_TRAJECTORIES_SNAPSHOT_JSON>

The snapshot contains every trajectory, including your own: each complete revised H, elite equations and their public errors, questions, analysis and experiment records, program-computed verdicts, evidence references, and unresolved issues.

Compare the scientific explanations, evidential support, prediction performance, and remaining uncertainty of all trajectories. You may vote for your own trajectory. Vote independently; do not infer or imitate another scientist's vote. Candidate import records do not change provenance.

Return:
- takeaway
- vote_for_trajectory
- vote_rationale
- import_candidate_ids
- imported_claim_or_evidence_refs
- next_open_question, containing claim_to_test, relation_to_own_conjecture,
observations_needed, and outcome_implications
- evidence_refs

Do not issue a merge command and do not replace the full H in this call. Do not count votes, decide termination, or select the final equation; those are program-side actions.
\end{promptblock}

\textbf{Output contract.} One JSON record per scientist. \texttt{vote\_for\_trajectory} may name any trajectory in the shared snapshot, including the voter's own.

\subsection{Seven-section scientific report writer}
The report presents the scientific questions, evidence obtained, and changes in interpretation during the investigation. It summarizes consequential tests and their evidential scope without reproducing model deliberation or individual tool calls.

The report writers for SCOPE, ReAct, and Reflexion-Exp receive the same exemplar and writer notes for report organization and expression; each method's evidence digest supplies its task-specific facts. Exemplar-guided report generation has also been used in ~\cite{yang2026multimodal}.

The questions in the seven-section prompt below illustrate possible coverage rather than a list requiring separate answers. The report follows the actual investigation and its evidence, without adding unsupported content to fill a section.

\role{System message}
\begin{promptblock}
Write the scientific report of a completed SCOPE closure investigation in the voice of a finished paper. Report what the evidence establishes and where it stops.

Do not expose chain-of-thought, private deliberation, hidden evaluation, or low-level tool logs. Do summarize the scientifically meaningful evidence development: the unresolved question, the evidence obtained, the registered prediction it supported or challenged within the specified window and criterion, and the resulting change or constraint on the scientific interpretation.
\end{promptblock}

\role{User message}
\begin{promptblock}
PUBLIC TASK
<PUBLIC_TASK_CARD>

PROGRAM-SELECTED TRAJECTORY AND FINAL EQUATION
<SELECTED_TRAJECTORY_JSON>
<PROGRAM_RENDERED_FINAL_EQUATION_BLOCK>

PUBLIC EVIDENCE DIGEST
<SCIENTIFIC_EVIDENCE_DIGEST_JSON>

ALTERNATIVE TRAJECTORIES
<ALTERNATIVE_EXPLANATIONS_JSON>

Write a coherent scientific report using the seven sections below. The questions describe the intended coverage of each section; organize the narrative around the investigation and its findings.

Use these display titles, in this order:
1. Problem and evidence scope
2. Observed phenomenology
3. Evidence development
4. Recovered closure and numerical behavior
5. Scientific findings and implications
6. Evidence boundary and numerical evaluation
7. Recommended discriminating experiments

Write exactly these seven sections, in this order, as fields of one JSON object. Each field is a list of prose statements that the program will join into paragraphs.

1. problem_and_scope -- What closure is unknown, in what variables, and what public
   record was available to investigate it?
2. phenomenology -- What did the system do before any candidate equation is named?
3. evidence_development -- Which scientific question motivated each consequential
   investigation, what evidence was obtained, which registered prediction was supported
   or challenged within its specified window and criterion, and what scientific
   interpretation changed? What remains unresolved?
4. closure_behavior -- What does each factor of the final equation do, and what do public
   property checks establish about its numerical behavior and scope?
5. findings -- What mechanisms or properties does the evidence support, what alternatives
   does it challenge, and what implications follow?
6. evidence_boundary -- Where does the evidence stop, including inconclusive or
   not_obtained tests, uncovered windows, instability, and alternatives not separated?
7. recommended_experiments -- Which discriminating experiments would resolve the open
   questions, and which evidence boundary would each address?

Write finished scientific prose, not a chronology of agent actions. Mention no internal IDs, case labels, checker names, or verdict labels in the body. Use numbers only when they carry the argument. Put limitations in evidence_boundary rather than appending a caveat to every statement. Retain competing explanations when the evidence does not separate them. Use only the supplied public digest; do not infer facts from hidden evaluation.

Each prose statement record contains text, type, evidence_ids, and scope. An equation_property statement also contains equation and check. An evidence-boundary statement also contains covers_groups.

Also return the machine-record fields:
- open_questions
- refutations_and_unobtained
- selection_rationale
- cited_candidate_ids
- cited_experiment_case_ids
- key_evidence_ids
\end{promptblock}

\section{Evaluation Metrics and Design Rationale}
\label{app:metric-rationale}

The main paper reports equation recovery quality and scientific discovery quality separately rather than collapsing them into a
single aggregate score. ID/OOD NMSE and accuracy-to-tolerance (Acc) directly evaluate the predictive error and generalization of
the final candidate equation and are necessary for establishing its numerical validity. Numerical accuracy alone, however, does
not reveal how much numerical improvement active experimentation achieves relative to the initial-evidence checkpoint, whether
the final report remains faithful to the discovery trajectory, or whether it identifies scientifically meaningful mechanistic
properties. Two candidate equations can
have similar numerical errors over the test domain while implying different directions of influence, limiting behaviors, or
variable couplings. Conversely, a method that has not yet recovered the complete closure relation may nevertheless have
identified correct and testable mechanisms that can guide subsequent experiments. ClosureBench therefore complements its
numerical equation metrics with three dimensions of interactive scientific discovery: experimental effectiveness, evidence
faithfulness, and mechanistic understanding.

\subsection{Equation-Recovery Metrics}
\label{app:equation-metrics}

For each task, we evaluate the final equation separately on its fixed
in-domain (ID) and out-of-domain (OOD) test sets. Let $n$ be the number of
test points, $y_j$ the reference output, and $\hat y_j$ the predicted output.
We report two complementary measures of numerical equation quality.

\paragraph{Normalized Mean Squared Error (NMSE).}
NMSE normalizes the mean squared prediction error by the variance of the
reference outputs:
\begin{equation}
\mathrm{NMSE}
=\frac{\sum_{j=1}^{n}(\hat y_j-y_j)^2}
       {\sum_{j=1}^{n}(y_j-\bar y)^2},
\end{equation}
where $\bar y=\frac{1}{n}\sum_{j=1}^{n}y_j$. NMSE uses all test points and
is defined when the reference variance is nonzero.

\paragraph{Accuracy-to-Tolerance (Acc).}
Following SR-Scientist~\citep{xia2025sr}, we allow the worst $5\%$ of
pointwise relative errors to exceed the tolerance.
Let $\mathcal I_{95}$ contain the $\lceil0.95n\rceil$ test points with the
smallest relative errors.
A task is solved if the remaining maximum error satisfies the tolerance:
\begin{equation}
\mathrm{Acc}_{\delta}
=\mathbf{1}\!\left\{
\max_{j\in\mathcal I_{95}}
\left|\frac{\hat y_j-y_j}{y_j}\right|\le\delta
\right\}.
\end{equation}
We use $\delta=0.01$ (Acc@1\%) and $\delta=0.001$ (Acc@0.1\%). For a single
task and test set, the score is $1$ if solved and $0$ otherwise.

\subsection{Scientific Discovery Quality}

\paragraph{Complete Symbolic Equivalence and Mechanistic Understanding.}
Symbolic Accuracy (SA), introduced by LLM-SRBench, uses a model-based evaluator to determine whether a predicted equation is
symbolically equivalent to the ground-truth equation after abstracting away specific parameters and constants. This avoids the
failure of exact string matching to recognize algebraically equivalent expressions or alternative program representations
~\cite{llmsrbench}. SA is well suited to answering whether a complete, symbolically equivalent equation has been recovered, but
it still treats the entire expression as a binary unit of evaluation. A prediction is not counted as successful if it omits a
necessary factor, adopts an approximation that is valid only over a limited domain, or fails to recover the complete coupling
structure. SA therefore cannot distinguish between obtaining no useful insight and recovering an incomplete equation that
nevertheless captures part of the underlying mechanism.

The results of SR-Scientist~\citep{xia2025sr} further illustrate the limited resolution of this evaluation granularity under
current model capabilities. SR-Scientist performs long-horizon, tool-augmented optimization and evaluates large-scale backbones including
GLM-4.5-Air, GPT-OSS-120B, and Qwen3-Coder-480B-A35B. Nevertheless, its highest reported SA on LSR-Synth is only $7.75\%$, achieved with
GLM-4.5-Air. The point is not that SA is itself inappropriate, but that complete symbolic recoveries remain
sparse even with large-scale models and long-horizon optimization, leaving little resolution for characterizing partial
mechanistic progress.

Based on this distinction, ClosureBench does not include SA in its evaluation suite. Candidate equations are already evaluated
using NMSE and Acc on fixed ID/OOD test sets. For the scientific report as the final discovery product, we instead ask whether
the method identifies scientifically meaningful variable dependencies, directions of influence, limiting behaviors, invariances,
and coupling relationships in the hidden closure relation. Mechanism Insight (MI) evaluates these properties against a
predefined and frozen set of reference mechanism facts and determines, at the level of complete sentences, whether the report
identifies
each mechanism completely, partially, or incorrectly. Different linguistic formulations may express the same mechanism, and
failure to recover exactly the same algebraic form as the ground-truth closure does not automatically imply an absence of
mechanistic understanding.

This design neither relaxes the requirements for equation correctness nor reclassifies approximate or incorrect equations as
correct. NMSE and Acc continue to evaluate the numerical accuracy and extrapolative performance of candidate equations, whereas
MI evaluates only the mechanistic understanding expressed in the report. The reference facts, their domains of applicability,
and the evaluation rubric are frozen before formal method outputs are examined. Method-blind sentence-level evaluation and
calibration against human annotations are then used to reduce the subjectivity inherent in open-ended language evaluation.
Evidence Faithfulness (EF) independently assesses whether mechanism statements are supported by the method's own trajectory;
consequently, post hoc guesses that are detached from discovery evidence cannot by themselves constitute high-quality scientific
discovery.

\paragraph{Relationship between EE and Numerical Equation Metrics.}
Final ID/OOD NMSE and Acc measure the absolute numerical quality attained by the equation at the end of discovery, whereas
Experimental Effectiveness (EE) measures the net improvement achieved by the same method over the course of active
experimentation. EE uses the change in the fraction of test points satisfying the 1\% relative-error tolerance, so gains and losses have a direct percentage-point interpretation. Across tasks, EE is arithmetically averaged over valid initial--final pairs. EE neither replaces final NMSE or Acc nor attributes an improvement to any single experiment; it measures only the overall numerical benefit of the complete active discovery process relative to the initial evidence.

\paragraph{Complementarity of the Metrics.}
The four metric families evaluate different objects. ID/OOD NMSE and Acc assess the absolute numerical quality of the final
equation; EE assesses the overall numerical improvement achieved by the active discovery process relative to the initial
evidence; MI assesses whether the final report identifies correct mechanistic properties; and EF assesses whether the report's
conclusions remain faithful to the evidence produced in the method's own trajectory. A high EE does not guarantee that the final
equation has reached a low error or that the report's mechanistic interpretation is correct. A high MI does not guarantee that
the corresponding insights are supported by the method's own evidence, and a high EF does not imply that the reported
mechanisms agree with the reference facts. Reporting these metrics separately distinguishes final equation accuracy, the benefit
of active experimentation, mechanistic understanding, and evidential support, while avoiding a single aggregate score that
obscures distinct failure modes.

\section{Human Validation of the LLM Evaluators}
\label{app:evaluator}

Evidence Faithfulness (EF) and Mechanism Insight (MI) require semantic
judgments over open-ended scientific text. For EF, a fixed, method-blind LLM
evaluator determines whether each report claim is fully supported by the
corresponding discovery-trace evidence. For MI, the evaluator determines the
degree to which a report claim matches a frozen reference mechanism fact,
assigning a score of \(1\), \(0.5\), or \(0\) for complete recovery, partial
recovery, or no recovery/contradiction, respectively. The LLM outputs only
these structured labels; the final EF and MI values are computed
deterministically. Experimental Effectiveness, NMSE, and Acc do not use an LLM
evaluator.

\begin{figure}[H]
  \centering
  \includegraphics[
    page=1,
    width=0.96\textwidth
  ]{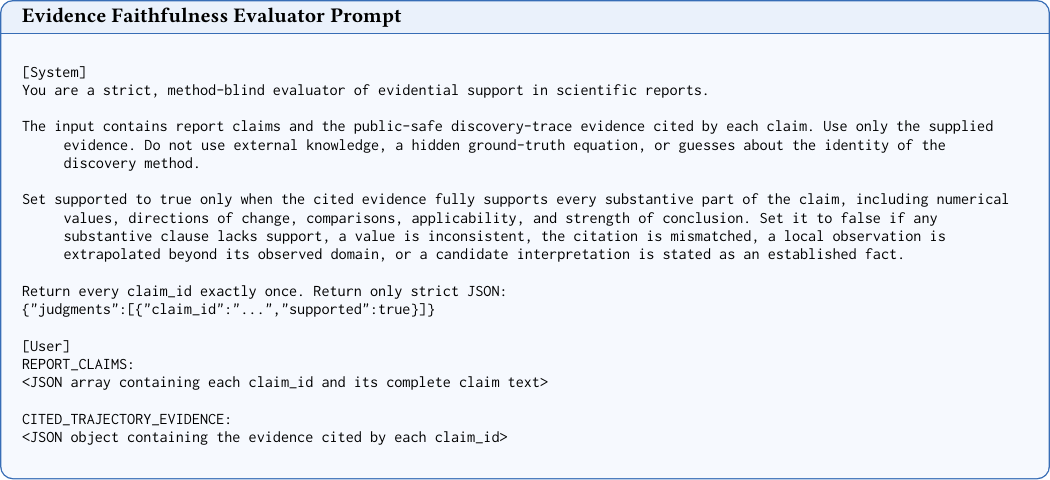}
  \caption{Complete prompt used by the method-blind Evidence Faithfulness
  evaluator. A full-width boxed rendering of the Evidence Faithfulness
  evaluator prompt, including its system instruction, strict JSON output
  schema, report-claim input, and cited trajectory-evidence input.}
  \label{fig:ef_evaluator_prompt}
\end{figure}

\begin{figure}[H]
  \centering
  \includegraphics[
    page=2,
    width=0.96\textwidth
  ]{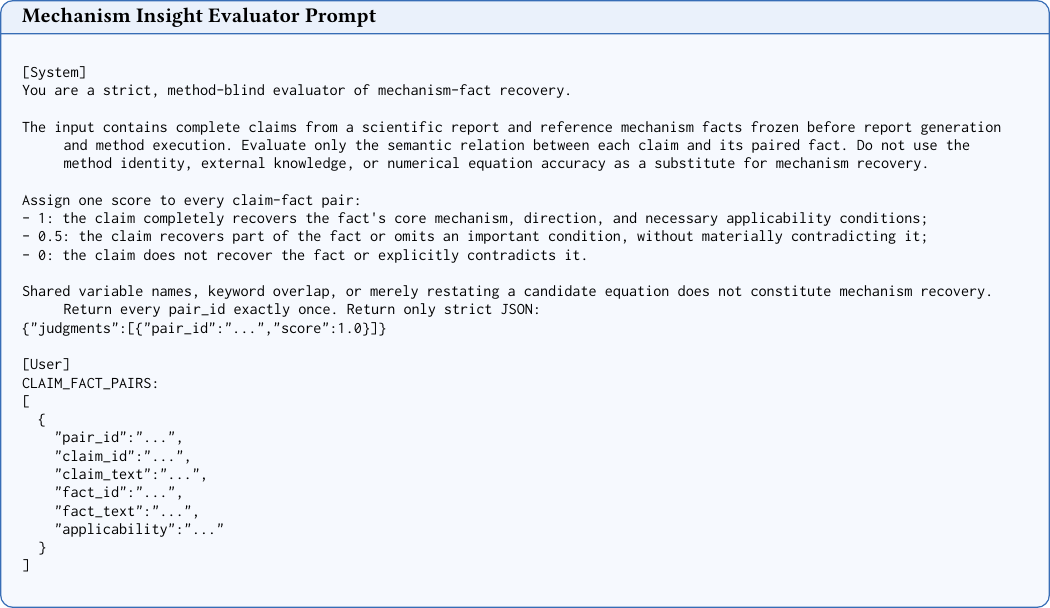}
  \caption{Complete prompt used by the method-blind Mechanism Insight
  evaluator. A full-width boxed rendering of the Mechanism Insight evaluator
  prompt, including its three-level claim--fact recovery rubric and strict JSON
  output schema.}
  \label{fig:mi_evaluator_prompt}
\end{figure}

\textbf{Evaluator configuration.~}
Following SR-Scientist's choice of evaluation model, we use GPT-OSS-120B to assess evidence faithfulness and mechanism insight. The evaluator returns
structured outputs that conform exactly to the schemas specified in
Figures~\ref{fig:ef_evaluator_prompt} and~\ref{fig:mi_evaluator_prompt}.
It is not provided with the identity of the method being evaluated and has
no access to external tools or the hidden ground-truth equations. For each
task, all methods are evaluated using identical prompt templates, frozen
reference facts, and inference settings.

\paragraph{Human validation.}
Following prior validation of model-based evaluators in
LLM-SRBench~\cite{llmsrbench} and SR-Scientist~\cite{xia2025sr}, two authors
independently assessed 200 EF claims and 150 MI claim--fact pairs sampled from
development reports. EF used the same binary supported/unsupported labels as
the automatic evaluator, whereas MI used the same \(1/0.5/0\) recovery scale.
The authors reconciled disagreements through discussion, after which the fixed
LLM evaluator assessed the same items. For EF, agreement requires an identical
binary label; for MI, it requires an identical recovery level.

\begin{table}[H]
  \centering
  \caption{Agreement between the LLM evaluator and human annotations.}
  \label{tab:llm_human_agreement}
  \small
  \begin{tabular}{lccc}
    \toprule
    Evaluation task & Items & Human--human & LLM--human \\
    \midrule
    EF evidence support
      & 200 & 94.5\% & 92.0\% \\
    MI fact matching and recovery
      & 150 & 92.7\% & 88.0\% \\
    \bottomrule
  \end{tabular}
\end{table}

As shown in Table~\ref{tab:llm_human_agreement}, the fixed evaluator closely
reproduces human judgments for both tasks. Agreement is lower for MI than for
EF because MI additionally distinguishes complete from partial mechanism
recovery. Most MI disagreements occur between adjacent recovery levels; no
entirely unrelated claim is assigned complete-recovery credit in the validation
sample. The complete evaluator prompts are reproduced below.

\FloatBarrier
\section{Paper-Anchored External Research Case: JFM Wake Closure}
\label{app:jfm-wake-report}

\definecolor{MICompleteFill}{HTML}{DDF2E1}
\definecolor{MIPartialFill}{HTML}{FFF0B8}
\definecolor{MIIncorrectFill}{HTML}{F8D4D1}
\definecolor{MIOnlyMELDFrame}{HTML}{287A46}
\soulregister\mbox1

\newif\ifMIShowAnnotations
\MIShowAnnotationstrue

\newcommand{\MIAnnotationsOn}{\MIShowAnnotationstrue}
\newcommand{\MIAnnotationsOff}{\MIShowAnnotationsfalse}

\newcommand{\MIHighlightedSentence}[2]{%
  \begingroup%
    \sethlcolor{#1}%
    \hl{#2}%
  \endgroup%
}

\newcommand{\MIPlainSentence}[1]{#1}

\newcommand{\MIComplete}[1]{%
  \ifMIShowAnnotations
    \MIHighlightedSentence{MICompleteFill}{#1}%
  \else\MIPlainSentence{#1}\fi}
\newcommand{\MIPartial}[1]{%
  \ifMIShowAnnotations
    \MIHighlightedSentence{MIPartialFill}{#1}%
  \else\MIPlainSentence{#1}\fi}
\newcommand{\MIIncorrect}[1]{%
  \ifMIShowAnnotations
    \MIHighlightedSentence{MIIncorrectFill}{#1}%
  \else\MIPlainSentence{#1}\fi}
\ulposdef{\MIOnlyMELDInlineDecoration}[xoffset=.08em]{%
  \begingroup
    \setlength{\fboxsep}{0pt}%
    \setlength{\fboxrule}{0.45pt}%
    \raisebox{-.55ex}[0pt][0pt]{%
      \fcolorbox{MIOnlyMELDFrame}{MICompleteFill}{%
        \rule{0pt}{2.65ex}%
        \rule{\dimexpr\ulwidth-2\fboxrule\relax}{0pt}}}%
  \endgroup}
\newcommand{\MIOnlyMELDComplete}[1]{%
  \ifMIShowAnnotations
    \unskip\nobreak\hspace{.25em}%
    \MIOnlyMELDInlineDecoration{#1}%
  \else\MIPlainSentence{#1}\fi}
\newcommand{\MILegendChip}[2]{%
  \begingroup\setlength{\fboxsep}{1.1pt}\colorbox{#1}{\strut #2}\endgroup}
\newcommand{\MIOnlyMELDLegendChip}[1]{%
  \begingroup
    \setlength{\fboxsep}{1.1pt}%
    \setlength{\fboxrule}{0.55pt}%
    \fcolorbox{MIOnlyMELDFrame}{MICompleteFill}{\strut #1}%
  \endgroup}

\newcommand{\MILegend}{%
  \begingroup\small
  \noindent
  \MILegendChip{MICompleteFill}{Complete mechanism match}\quad
  \MILegendChip{MIPartialFill}{Partial mechanism match}\quad
  \MILegendChip{MIIncorrectFill}{Contradiction}\par\noindent
  \endgroup}

\newcommand{\MIOnlyMELDLegend}{%
  \begingroup\small
  \noindent
  \MIOnlyMELDLegendChip{Complete mechanism match}\quad
  Dark-green outline: fully resolved only by MELD in this Case 1 comparison; repeated outlined sentences still count once per fact.
  \par
  \endgroup}

\definecolor{ScientificReportFrame}{HTML}{376DB6}
\definecolor{ScientificReportBackground}{HTML}{F6F9FE}
\definecolor{ScientificReportTitleBackground}{HTML}{EAF1FB}

\newtcolorbox{ScientificReportBox}[1]{%
  enhanced,
  breakable,
  colback=ScientificReportBackground,
  colframe=ScientificReportFrame,
  colbacktitle=ScientificReportTitleBackground,
  coltitle=black,
  coltext=black,
  fonttitle=\bfseries\normalsize,
  fontupper=\fontsize{7.8}{8.7}\selectfont,
  title={#1},
  boxrule=0.65pt,
  arc=2mm,
  left=2.5mm,
  right=2.5mm,
  top=2mm,
  bottom=2mm,
  before skip=1.2em,
  after skip=1.5em,
  pad at break*=1mm,
  segmentation hidden
}

\subsection{Motivation and Experimental Setting}
\label{app:jfm-case-setting}

ClosureBench remains the main controlled benchmark. We additionally use a wind-turbine wake problem from Wang et al.~\citep{wang2025symbolic} as an external scientific research case. The purpose is to examine whether SCOPE can investigate a physical question grounded in a published study and produce scientific understanding supported by an accurate model and traceable evidence. Equation recovery is one stage of this investigation; the final report integrates the model, scientific conclusions, supporting evidence, limits of applicability, and unresolved questions.

\cite{wang2025symbolic}'s study is suitable because it provides both a quantitative reference and scientific properties against which the report can be assessed. Equation~(2.15) and the parameters in its Figure~7 define the reference force-density model, while the paper discusses mirror symmetry, two off-axis peaks, a central trough, downstream decay, and approximate self-similarity. Here \(\hat{x}=x/D\) and \(\hat{y}=y/D\), where \(D\) is the rotor diameter, and positive \(q_T\) denotes removal of axial momentum from the flow. Observations for this case are generated from the published reference model, but its expression and parameters are not given to SCOPE. SCOPE selects spatial coordinates \(\hat{x}\in[0,0.2]\) and \(\hat{y}\in[-1,1]\) at which to query observations. \uline{During discovery, literature retrieval is limited to publications preceding the target work}~\citep{wang2025symbolic}. The case begins with 1,000 observations and uses 3 scientists for at most 10 rounds and 30 batches of active queries. After the investigation, the published reference is used to assess the recovered equation and the scientific conclusions in the report.

This case uses GPT-5.5. We do not use GPT-5.5 for the ClosureBench benchmark because of its higher token cost.

\subsection{Scientific Report}
\label{app:jfm-annotated-report}

\begin{ScientificReportBox}{SCOPE Report on a JFM Wind-Turbine Wake Case}
\textbf{1. Problem and evidence scope.}
The task is to recover an interpretable closure for the dimensionless local axial turbine-force density \(q_T\) from the normalized downstream coordinate \(\hat{x}\in[0,0.2]\) and transverse coordinate \(\hat{y}\in[-1,1]\). Three scientists start from 1,000 initial observations, conduct 10 rounds of active experiments, and share the newly acquired observations. A literature snapshot predating the target paper and the initial data analysis supply the initial conjectures; subsequent analyses and experiments test and revise them.

\textbf{2. Observed phenomenology.}
\MIComplete{Across overlapping downstream windows, the positive and negative transverse profiles are consistent with an approximately mirror-symmetric force distribution. The observations contain two off-axis maxima of similar magnitude rather than a single centerline maximum. The centerline force lies below the off-axis maxima, forming a central trough. Within the observed interval, the peak force decreases as the downstream coordinate increases.} The initial coverage is uneven: evidence at low \(\hat{x}\) samples only the positive lobe, while most high-\(\hat{x}\) windows are one-sided and have different transverse spans.

\textbf{3. Evidence development.}
The 3 scientists begin respectively from center enhancement, side-concentrated loading, and relative center weakening within a broad envelope, exploring centered Gaussian or super-Gaussian, shifted double-Gaussian, and difference-of-Gaussians representations. Each research round first uses modeling and analysis to identify a question, then obtains evidence through a problem-solving experiment or tests an existing conclusion through a validation experiment; after every 2 rounds, a roundtable compares the conjectures, equations, evidence, and remaining disagreements.

Early experiments first test the centered single-peak conjecture. \MIComplete{The centerline-to-negative-lobe contrast is inconsistent with the centered single-peak prediction, and a later validation again observes the central trough, further weakening the centered Gaussian or super-Gaussian explanation. The conjecture is therefore revised to a double-peaked shape.} The downstream evolution is then treated mainly as a change in overall amplitude rather than another adjustment of a centered single peak.

After the double-peaked structure receives initial support, the investigation turns to downstream evolution and the stability of the transverse shape. The negative-side experiment at high \(\hat{x}\) mainly constrains decay but omits the centerline; the outer-shoulder experiment adds tail information but does not cover the peaks. A matched centerline, peak-region, and tail comparison then fills these gaps and tests the downstream weakening of amplitude. \MIComplete{The combined results support an approximately stable peak location and downstream change dominated by decreasing amplitude.} \MIPartial{The transverse-shape variation is comparatively weak, but the observations are insufficient to establish constant width or strict spatial separability.}

As the main shape features converge, the question becomes whether the different double-peaked representations are distinguishable. Experiments targeting center curvature and the outer shoulder compare the shifted double-Gaussian and difference-of-Gaussians forms. Both explain the observed structure in the main observation region, while equation analysis reveals strong cancellation between the terms in the difference of Gaussians. The investigation therefore shifts from comparing fit alone to asking whether the representation is reliably identifiable from the evidence and whether data coverage limits its range of validity.

\MIComplete{The final review further supports the double-peaked structure, approximately stable peak location, and decreasing downstream amplitude}, but the experiments do not eliminate the shifted double Gaussian and the edge observations do not determine a reliable extrapolation law. The fifth roundtable does not reach unanimity, and the investigation stops when the experiment budget is exhausted. The difference-of-Gaussians trajectory receives the most votes, and its elite equation with the lowest prediction error on the final public data is returned, while the shifted double Gaussian remains an alternative explanation compatible with the evidence.

\textbf{4. Recovered closure and numerical behavior.}
The locked candidate is
\[
\widehat q_T(\hat{x},\hat{y})=A(\hat{x})\exp\!\left[-\frac{\hat{y}^2}{2V(\hat{x})}\right]-B\exp\!\left[-\frac{\hat{y}^2}{2s^2}\right],
\]
with
\[
A(\hat{x})=501.5404(\hat{x}+28.2491)^{-0.195963},\quad
V(\hat{x})=0.2415296(\hat{x}+28.2491)^{-0.466942},
\]
and \(B=260.1523\), \(s=0.224824\). Here \(V\) is the width parameter of the broader Gaussian component. Both components are individually large and nearly cancel, so their coefficients cannot be interpreted as independent physical processes from the fit alone. \MIComplete{The expression is strictly even in the transverse coordinate and preserves a central trough and two symmetric off-axis peaks. Over the downstream interval from 0.01 to 0.20, the modeled peak position changes by approximately \mbox{$5.05\times10^{-4}D$}, and the peak envelope is approximately linear, with \mbox{$R^2>0.99999$} for its linear projection.} \MIPartial{The center-to-peak ratio and outer half-maximum location vary little, consistent with approximate shape preservation.} These are computed properties of the expression; they do not mean that the experiments resolve such a small peak-position change or prove that the physical distribution is strictly symmetric or separable.

\textbf{5. Scientific findings and implications.}
The evidence supports a two-lobed shape whose amplitude weakens clearly downstream while the peak locations and principal transverse profile vary more weakly. \MIPartial{The off-axis lobes are compatible with an explanation based on blade loading that peaks near mid-span, but the residual field does not independently establish this cause.} Over the finite domain, both the difference of Gaussians and the shifted double Gaussian describe the double-peaked form, so low global error is insufficient to identify a unique representation; the negative term in the difference does not imply an independent reverse force. The robust scientific conclusions are the two-lobed structure and its downstream decay, not separate physical counterparts for each component of the difference of Gaussians. The recovered expression is therefore a compact representation of the observed behavior rather than a confirmed unique generative mechanism.

\textbf{6. Evidence boundary and numerical evaluation.}
The final expression has complexity 29 and NMSE \(2.1086\times10^{-4}\). The difference of Gaussians and shifted double Gaussian remain unresolved in the principal observation region, so the low overall error does not identify a unique representation. The observations support an approximately stable peak location but do not establish constant width or strict spatial separability. \MIComplete{The force is concentrated immediately downstream of the rotor and decays to a negligible level by approximately 0.3 rotor diameters.} The current candidate only approaches zero near the edge of the registered support and does not explicitly encode non-negative truncation outside it. Low-signal edge observations do not determine the effective support or a reliable extrapolation law. The causal connection between the double-peaked structure and blade mid-span loading, and applicability under changing inflow turbulence, rotor operating state, and yaw, require additional physical experiments.

\textbf{7. Recommended discriminating experiments.}
First increase the precision of center-curvature and peak-to-shoulder measurements in matched profiles to separate the two remaining representations, then add low-signal edge observations and inspect residual error to determine the effective support. Add blade-resolved or radial-load evidence, followed by tests under different inflow-turbulence conditions, to distinguish the within-domain shape law, the blade-level cause, and cross-condition applicability.
\end{ScientificReportBox}

\subsection{Evaluation Against the Published Study}
\label{app:jfm-case-evaluation}

For quantitative comparison, Eq.~(2.15) of Wang et al.~\citep{wang2025symbolic} gives
\[
q_T^{\mathrm{JFM}}(\hat{x},\hat{y})=a(\hat{x})\left[
\exp\!\left(-\frac{(\hat{y}-\mu)^2}{2\sigma^2}\right)+
\exp\!\left(-\frac{(\hat{y}+\mu)^2}{2\sigma^2}\right)
\right],
\]
with \(a(\hat{x})=-2.33\hat{x}+0.57\), \(\mu=0.26\), and \(\sigma=0.19\). The locked candidate is algebraically different from this shifted double Gaussian and therefore does not recover its strict expression, while retaining ID NMSE \(2.1086\times10^{-4}\) and OOD NMSE \(4.1932\times10^{-4}\).

The report contains 11 deduplicated mechanism claims and the paper-anchored reference contains 10 facts. Seven facts match completely and 3 partially; the additional claim that a difference of Gaussians can represent the observed twin-lobed pattern is unmatched but not contradictory. The maximum one-to-one matching weight is \(W=8.5\), giving
\[
\mathrm{MI}=\frac{2W}{|\mathcal C|+|\mathcal R|}
=\frac{2\times8.5}{11+10}=\frac{17}{21}=0.8095.
\]

\begin{table}[H]
\centering
\caption{Paper-anchored mechanism facts used to evaluate the JFM scientific report.}
\label{tab:jfm-mi-reference-facts}
\small
\renewcommand{\arraystretch}{1.16}
\begin{tabular}{@{}>{\raggedright\arraybackslash}p{0.055\linewidth}>{\raggedright\arraybackslash}p{0.49\linewidth}>{\raggedright\arraybackslash}p{0.115\linewidth}>{\raggedright\arraybackslash}p{0.25\linewidth}@{}}
\toprule
ID & Paper-anchored mechanism fact & Judgment & Location in the official JFM PDF\\
\midrule
F01 & Mirror symmetry in the signed cross-wake coordinate. & \MILegendChip{MICompleteFill}{Complete} & \S2.3.2, (2.15), PDF p.~10\\
F02 & Two comparable off-axis maxima in the transverse force profile. & \MILegendChip{MICompleteFill}{Complete} & \S2.3.2, (2.15), fig.~6(b), PDF pp.~10--11\\
F03 & A centerline response below the off-axis maxima, producing a central trough. & \MILegendChip{MICompleteFill}{Complete} & \S2.3.2, fig.~6(b), PDF p.~11\\
F04 & An approximately constant off-axis peak location within the localized forcing region. & \MILegendChip{MICompleteFill}{Complete} & Fig.~7(d), \S3.1, PDF pp.~12--13\\
F05 & An approximately constant effective transverse lobe width. & \MILegendChip{MIPartialFill}{Partial} & Fig.~7(d), \S3.1, PDF pp.~12--13\\
F06 & Monotonic downstream weakening of force amplitude through the active interval. & \MILegendChip{MICompleteFill}{Complete} & Fig.~7(c), \S3.1, PDF pp.~12--13\\
F07 & Approximately linear downstream decay of force amplitude. & \MILegendChip{MICompleteFill}{Complete} & Fig.~7(c), \S3.1, PDF pp.~12--13\\
F08 & An approximately self-similar transverse pattern while amplitude weakens. & \MILegendChip{MIPartialFill}{Partial} & \S2.3.2 and \S3.1, PDF pp.~10 and 13\\
F09 & Forcing localized immediately downstream and negligible by approximately \(0.3D\). & \MILegendChip{MICompleteFill}{Complete} & \S2.3.2, PDF p.~10\\
F10 & A link between the off-axis structure and blade aerodynamics, particularly lift peaking near blade mid-span. & \MILegendChip{MIPartialFill}{Partial} & \S3.1, PDF p.~14\\
\bottomrule
\end{tabular}
\end{table}
\FloatBarrier

\clearpage
\section{Scientific Reports for Flow-Induced Crystallization}
\label{app:fic-scientific-reports}

\ifdefined\MIComplete\else
  
\fi
\ifcsname ScientificReportBox\endcsname\else
  \definecolor{ScientificReportFrame}{HTML}{376DB6}
\definecolor{ScientificReportBackground}{HTML}{F6F9FE}
\definecolor{ScientificReportTitleBackground}{HTML}{EAF1FB}

\newtcolorbox{ScientificReportBox}[1]{%
  enhanced,
  breakable,
  colback=ScientificReportBackground,
  colframe=ScientificReportFrame,
  colbacktitle=ScientificReportTitleBackground,
  coltitle=black,
  coltext=black,
  fonttitle=\bfseries\normalsize,
  fontupper=\fontsize{7.8}{8.7}\selectfont,
  title={#1},
  boxrule=0.65pt,
  arc=2mm,
  left=2.5mm,
  right=2.5mm,
  top=2mm,
  bottom=2mm,
  before skip=1.2em,
  after skip=1.5em,
  pad at break*=1mm,
  segmentation hidden
}

\fi

This appendix compares the ReAct, Reflexion-Exp, and SCOPE reports for the flow-induced crystallization task \texttt{fic\_task\_y}. The unknown local crystallization rate \(R_\psi(\psi,T,g)\) depends on crystallinity \(\psi\), absolute temperature \(T\), and normalized shear rate \(g\). It is evaluated from the local state at the beginning of each physical time step, acts as a positive source in the crystallinity equation, and enters the temperature equation through the known 18 K latent-heat temperature-rise coefficient. All three methods receive the same public task description and observation budget.

\subsection{Comparison of the Three Reports}

ReAct identifies shear promotion, a non-monotone temperature response, and several qualitative properties of the crystallinity response, but it interprets coupling in its selected expression as coupling in the crystallization process. Reflexion-Exp likewise recovers the main response directions and the seed-dependent onset, while making claims inconsistent with the reference mechanisms that shear changes the thermal-window shape and that crystallinity and shear are inseparable. SCOPE gives the most complete account of the ground-truth mechanisms among the three displayed reports: it identifies the competition between nucleation growth and depletion, the bounded temperature window, the separable input effects, the finite quiescent shear response, and the spatial confounding between shear and temperature. Its dark-green outlines mark the three complete mechanisms not fully stated by either displayed baseline: the spatial-confounding analysis, the coverage that brackets the temperature optimum, and the square-root shear factor with a finite zero-shear baseline.

Green, yellow, and red backgrounds denote complete mechanism matches, partial matches, and explicit contradictions. A dark-green outline marks a complete mechanism match stated only by SCOPE in this three-report comparison.

\subsection{Complete Scientific Reports}

\begin{ScientificReportBox}{ReAct: Flow-Induced Crystallization Scientific Report}
\textbf{1. Problem and evidence scope.}
The unknown closure is the local crystallization rate \(R_\psi(\psi,T,g)\), where \(\psi\) is local crystallinity, \(T\) is absolute temperature, and \(g\) is normalized shear rate. The rate is computed from the local state at the beginning of each physical time step and held fixed during that step; it enters the crystallinity equation as a positive source and the temperature equation through the known 18 K latent-heat temperature-rise coefficient. The public task description defines the physical constraints and variable ranges, including the 320 K glass-transition temperature and 440 K melting point, which provide state boundaries and physical references for constructing the closure. The record contains 30 cases and 4,000 rows. The investigation begins from an initial closure representation and progressively extends the evidence coverage to test whether that representation can jointly explain shear promotion, the temperature window, and changes along the crystallinity direction. Additional simulation observations are used mainly to assess response regions not covered by the initial explanation. Literature retrieval provides qualitative background on the temperature window, shear promotion, and material conversion, while the retained response structure is determined by the observations.

\textbf{2. Observed phenomenology.}
\MIPartial{Across the binned operating conditions, the crystallization rate increases monotonically with normalized shear rate, with no internal extremum in the sampled range. The temperature response first rises, reaches an internal maximum near the cold end of the sampled interval, and then declines toward the hot end.} \MIComplete{The crystallization-rate response to crystallinity first rises and then falls, forming a single internal maximum.} \MIPartial{The temperature turning point is comparatively well constrained across the operating conditions.} \MIComplete{Completely amorphous material requires an existing crystal seed for crystallization to begin.}

\textbf{3. Evidence development.}
The investigation examines how the initial closure represents \MIPartial{shear enhancement}, \MIPartial{an internal maximum along the temperature direction}, and \MIComplete{the onset mechanism in which an existing crystal population promotes early conversion and the rate has a square-root dependence on crystallinity}. Candidate comparisons show prediction bias in the initial representation; the limitation cannot be removed by parameter adjustment alone but reflects an inadequate response structure, and its relative error is 89.8\% higher than that of the retained form. The subsequently explored candidates provide different response structures. Among 102 candidates, 3 have similar predictive performance, and the current representation is selected by complexity. A complete closure must jointly represent \MIPartial{monotone increase along the shear direction}, \MIPartial{an internal turning point along the temperature direction}, and \MIComplete{a temperature response that vanishes at the glass-transition and melting endpoints and is positive inside the window}, rather than improve only one response direction. The remaining uncertainty concerns identifiability among approximate candidate representations. A candidate with a different temperature variable gives nearly identical predictions, so the evidence supports structural revision of the initial representation but does not establish a unique approximate temperature function.

\textbf{4. Recovered closure and numerical behavior.}
The selected expression for the current observations is
\[
R_\psi=a\bigl((\cos(\log(b\psi))+g^c+\psi)e^{\sqrt{\psi}}\bigr)^{d^{\sin(g)}}\cos(k\sqrt{T}),
\]
where \(a=0.007104\), \(b=0.00997\), \(c=0.7421\), \(d=1.011\), and \(k=0.9682\). \MIComplete{The temperature response is a symmetric parabolic curve about the midpoint of the window, so equal temperature offsets on the cold and hot sides produce equal rate decreases.} The candidate's cosine temperature term is an approximation over the sampled range. \MIIncorrect{Shear promotion and crystallinity are inseparable, so the same shear change produces different multiplicative rate increases at different crystallinities.} This interpretation follows from the candidate structure, in which the shear and crystallinity terms enter the same parentheses. \MIPartial{The cosine temperature factor produces a non-monotone window consistent with the binned results over the covered range and an internal maximum near the cold end.} \MIPartial{The candidate represents shear promotion and an internal temperature maximum, but several approximate expressions remain difficult to distinguish.} Fixing shear rate, temperature, and crystallinity in turn at their medians changes the prediction by 20.7\%, 19.7\%, and 8.7\% of its own scale, respectively, so shear rate and temperature are the dominant inputs. Another retained candidate and the final selected equation have prediction correlation 0.99996 and differ in relative error by 3.2\%, limiting conclusions about the specific temperature-function form.

\textbf{5. Scientific findings and implications.}
The observations support \MIPartial{shear promotion}, a \MIPartial{non-monotone temperature response}, and \MIComplete{a conversion process in which nucleation growth competes with loss of unconverted material and late behavior is depletion dominated}. The current expression compresses these features into a computable quantitative relation, but its algebraic terms do not separately correspond to identified microscopic processes. \MIIncorrect{For the same increment in crystallinity, the increase in rate relative to its previous value also changes with the shear level.} \MIComplete{Restricted segmental motion at the cold end and crystal destabilization at the hot end jointly form the crystallization-temperature window.} \MIPartial{As required by the physical boundaries in the task, the temperature window is bounded by the 320 K glass-transition temperature and the 440 K melting point}; however, the current candidate remains an approximation over the sampled range and its particular algebraic representation is not uniquely determined.

\textbf{6. Evidence boundary and numerical evaluation.}
The current and alternative expressions remain indistinguishable on the available record. The analysis identifies conditions that could amplify their prediction differences, but observations at those conditions have not been obtained. Low error therefore does not close the identifiability gap. Refitting under the available operating conditions shows that constants \(a\) and \(b\) vary with the initial crystallinity seed, limiting their interpretation as universal constants in one closure law. Cosine, Gaussian, and other approximations remain difficult to distinguish over the sampled range. Agreement between the fit and the binned trends cannot replace tests at uncovered boundaries or conditional responses with the other variables fixed.

\textbf{7. Recommended discriminating experiments.}
First obtain conditions that maximize the prediction difference between the current and alternative representations, then compare the directions and magnitudes of their residuals. Only persistent support for one form together with systematic bias in the other can narrow the identifiability gap. In a series with fixed inlet velocity and wall temperature, vary only the initial crystallinity seed and compare records at similar local states to determine whether parameter changes reflect genuine condition dependence or compensation induced by the limited design. Increase wall-temperature resolution on both sides of the internal temperature maximum, focusing on peak curvature and the decline on each side, to distinguish the square-root cosine window from other non-monotone candidates.
\end{ScientificReportBox}

\begin{ScientificReportBox}{Reflexion-Exp: Flow-Induced Crystallization Scientific Report}
\textbf{1. Problem and evidence scope.}
The physical closure to be recovered describes the local crystallization rate. It is a positive algebraic source computed at the beginning of each physical time step from local crystallinity, absolute temperature, and normalized shear rate and held fixed during that step. It drives the crystallinity equation and feeds back to the temperature equation through the known 18 K latent-heat coefficient. The task card gives the allowed ranges of the three inputs and the material-transition constraints, including the 320 K glass-transition temperature and 440 K melting point. The record contains 30 cases and 4,000 rows, with 100 rows per case containing the three inputs and the observed rate. The investigation begins from initial observations and an existing closure representation, then adjusts its understanding of the response structure when later evidence exposes systematic bias. Additional simulation observations test regions not covered by the early explanation rather than merely enlarging the dataset.

\textbf{2. Observed phenomenology.}
\MIPartial{The marginal temperature response has a turning point: the rate rises from the cold end of the sampled range, peaks in the lower-middle part of the window, and then declines toward the hot end.} \MIPartial{The marginal response to shear rate increases monotonically throughout the sampled range.} \MIComplete{The crystallization-rate response to crystallinity is jointly governed by early nucleation enhancement and later depletion of unconverted material, rising and then falling to form a single internal maximum; the initial seed lies on the rising branch.} \MIIncorrect{Temperature and shear have the strongest interaction in the expression: between the lower and upper halves of the shear range, the rank correlation of the temperature response changes by 0.63, closely matching the observed change of 0.64.}

\textbf{3. Evidence development.}
The research starts from 8 initialized forms, all explicit polynomial or power-law expressions that couple the factors \(T-T_g\) and \(T_m-T\) to different powers of shear and crystallinity. A search over 233 candidates yields a selection band of 7 candidates. Although the locked form's public error is 0.000532, slightly above the leading band's 0.000471, it is selected because it has the lowest complexity in the band, 24. On the available rows, a band-leading form with logarithmic temperature dependence and different shear coupling is indistinguishable from the locked form: their prediction correlation is 0.9995 and their relative errors differ by 1.1\%. The initial polynomial-window form is distinguishable from the locked form, with a 79.3\% relative-error difference. Parameter-stability checks show that all 5 constants of the locked form stay within 20\% of their respective medians across 8 conditions, whereas 2 of the 4 constants in the initial form drift with wall temperature.

\textbf{4. Recovered closure and numerical behavior.}
\[
R_\psi=a e^\psi\bigl(b-\cos(-\sqrt{T}/c)\bigr)^d\sqrt{k e^\psi+g},
\]
where \(a=0.01328\), \(b=1.102\), \(c=0.4134\), \(d=0.3119\), and \(k=0.3281\). The exponential crystallinity factor contributes monotone increase, while the square-root factor couples crystallinity and shear sublinearly through \(k=0.3281\). \MIPartial{The cosine factor supplies a non-monotone temperature window: its frequency is set by c = 0.4134, while the sublinear exponent d = 0.3119 compresses the window so that the rate peaks in the lower-middle part of the sampled range, away from the cold and hot boundaries.} \MIPartial{The candidate represents shear promotion and an internal temperature maximum, but several approximate expressions remain difficult to distinguish.} Fixing shear rate at its median changes the prediction by 20.8\% of its own scale; fixing temperature gives 19.0\%, and fixing crystallinity gives 8.1\%, so shear and temperature are the dominant inputs. All 5 constants remain stable across 8 conditions without drifting toward any control variable. The band-leading form is indistinguishable from the locked form on the sampled rows, so the choice is based on complexity rather than a decisive error advantage.

\textbf{5. Scientific findings and implications.}
The locked form approximates crystallization mechanisms through several terms: \MIPartial{the cosine factor implements a non-monotone temperature window}, \MIPartial{the square-root factor implements shear promotion}, and \MIComplete{a completely amorphous state cannot initiate crystallization by itself; an existing crystal seed promotes conversion, with square-root crystallinity dependence at onset.} \MIComplete{The temperature response vanishes at the glass-transition and melting endpoints and is positive inside the window.} The locked form does not exactly preserve the onset, endpoint, and non-monotone properties of the crystallization process. The mixed term is interpreted as implying that \MIIncorrect{the effects of crystallinity and shear are inseparable, so the same shear change produces different relative rate increases at different crystallinities.} \MIIncorrect{A strong temperature-shear interaction means that the temperature window becomes flatter at high shear and steeper at low shear, so its effective width depends on local shear rate.} The selected form is 1 of 7 forms in the selection band and cannot be uniquely identified from the available rows.

\textbf{6. Evidence boundary and numerical evaluation.}
The locked form has public NMSE 0.000532 at complexity 24. The band-leading form is indistinguishable from it: prediction correlation is 0.9995 and relative error differs by 1.1\%. The model predicts 3 conditions that would distinguish the forms, but none was acquired. The sampled temperature range is 347.7 K to 424.9 K, entirely within the 320 K to 440 K window but reaching neither boundary, so the cosine window's edge shape is unconstrained. A planned middle-region condition with moderate inlet velocity, crystallinity seed, and wall temperature duplicated an existing condition, leaving the central-window behavior at moderate shear without independent confirmation.

\textbf{7. Recommended discriminating experiments.}
Acquire the predicted conditions that separate the locked and band-leading forms to determine which algebraic structure is supported where the present 4,000 rows cannot distinguish them. Acquire the previously duplicated middle-region condition with moderate inlet velocity, crystallinity seed, and wall temperature to confirm the central-window behavior independently at moderate shear. Extend wall-temperature control toward the cold end, closer to 320 K, to test whether the cosine-window shape remains valid at the currently unreached boundary and to constrain \(c\).
\end{ScientificReportBox}

\begin{ScientificReportBox}{SCOPE: Flow-Induced Crystallization Scientific Report}
\textbf{1. Problem and evidence scope.}
The study concerns a local crystallization rate whose inputs are crystallinity, absolute temperature, and normalized shear rate. Its positive sign denotes a source in the crystallinity balance, and it also enters the temperature equation through the 18 K latent-heat coefficient. The public task card specifies the three input variables and their allowed ranges, together with inlet velocity, crystallinity seed, and wall temperature as experimental controls, but specifies no functional form, threshold, or saturation scale. Three scientists share 1,000 initial observations and conduct the investigation within a budget of 30 additional simulation experiments. Three conjecture families are initialized from prior physical reasoning about bounded saturation, Gaussian-Avrami depletion, and separable power-law thermal scaling.

\textbf{2. Observed phenomenology.}
Across the sampled window, increasing shear is compatible with increasing crystallization rate; in the region covered by high inlet velocity, shear rate and rate remain clearly positively correlated. This supports shear promotion in the covered region but does not by itself determine its curvature or establish a saturation plateau at higher shear. An important temperature constraint is the decline at the hot end. At high crystallinity, the crystallization rate weakens markedly as temperature increases. The crystallinity direction cannot be summarized by simple monotone depletion: \MIPartial{where high crystallinity overlaps high temperature, crystallinity and rate remain positively correlated.}

\textbf{3. Evidence development.}
The stage change in crystallization is that \MIComplete{nucleation growth competes with consumption of unconverted material, producing a transition from pre-peak enhancement to post-peak weakening.} The relative RMSE is 0.0043 for the non-monotone candidate and 0.0103 for the monotone form. Under high crystallinity and high temperature, the \MIPartial{observed correlation is strongly positive}. Over the same state interval, the rate declines sharply with temperature. In data acquired at high velocity, shear and rate remain strongly positively correlated. Separating the input effects also requires their spatial distribution: \MIOnlyMELDComplete{in a fixed shear field, the wall region has both stronger shear and lower temperature, while the centerline has no shear; these effects are spatially confounded within one condition, wall-temperature contrasts separate them, and centerline states identify the quiescent crystallization baseline.} \MIPartial{The temperature window remains bounded}; \MIComplete{the temperature response vanishes at the glass-transition and melting endpoints and is positive inside the window.} Different conditions also open different temperature ranges: \MIOnlyMELDComplete{colder-wall conditions span both sides of the optimum, whereas the warmest-wall conditions cover only the descending hot side, so the optimum is constrained mainly by the colder conditions.} The candidate's specific approximate form is still not uniquely determined.

\textbf{4. Recovered closure and numerical behavior.}
\[
R_\psi=\sqrt{g+a}\left[b\sin(\psi^c-\psi^2)(T-d)(k-T)-T^m\right],
\]
where \(a=0.4\), \(b=1.384\times10^{-5}\), \(c=0.4709\), \(d=319.4\), \(k=440.3\), and \(m=-1.398\). \MIOnlyMELDComplete{The square-root shear factor with offset a = 0.4 gives sublinear promotion and remains finite at zero shear.} Fixing shear at its median changes the prediction by 27\% of its own scale, the largest of the three inputs, followed by temperature at 18.2\% and crystallinity at 17.6\%. \MIPartial{With d = 319.4 and k = 440.3, the product (T - d)(k - T) forms a parabolic temperature window that is zero at both endpoints and peaks in the interior.} \MIComplete{The effects of the inputs on crystallization rate can be multiplied separately; shear changes the overall magnitude without moving the temperature or crystallinity optimum.} The current candidate is only an approximation over the sampled range and does not preserve the separable structure and limiting properties of the crystallization rate term by term. For the candidate itself, the sine factor with \(c=0.4709\) is algebraically non-monotone but \MIPartial{increases monotonically over the sampled crystallinity range}, producing a monotonically increasing binned mean while retaining the ability to turn at higher crystallinity. The subtracted \(T^m\) term increases as temperature decreases when \(m=-1.398\), reducing the net rate at the cold end and making the lower temperature boundary steeper. Across 30 conditions, all 6 constants of the locked expression remain within 20\% of their respective medians. Constants in the initial expression drift with crystallinity seed, confirming that it is a condition-dependent fit rather than one law. The error ratio between the locked and initial expressions is 45, but the locked expression cannot be distinguished from the runner-up: prediction correlation is 1.0 and relative error differs by less than 1\%.

\textbf{5. Scientific findings and implications.}
\MIComplete{The existing crystal population determines onset: completely amorphous material cannot initiate crystallization by itself, and the initial rate grows with existing crystallinity in square-root form}, although the record does not isolate the underlying microstructure. The closure is local and history independent: the rate depends only on the instantaneous values of the three inputs, not on the crystallinity seed.

\textbf{6. Evidence boundary and numerical evaluation.}
The locked expression achieves public NMSE \(7.16\times10^{-7}\) at complexity 32 and is selected from a 2-form band among 701 candidates. The runner-up remains indistinguishable on the sampled rows, with less than 1\% relative-error difference. A separable power-law temperature form is supported in some windows, but the candidate's specific approximation is not uniquely determined. The sine crystallinity factor is constrained only over the sampled range; its algebraic form predicts a turn at higher crystallinity, but no acquisition reaches that state interval.

\textbf{7. Recommended discriminating experiments.}
Acquire the 3 identified conditions that separate the locked form from the runner-up, placing the flow at specific combinations of inlet velocity, crystallinity seed, and wall temperature where the 2 algebraic forms diverge. Acquire a low-temperature, low-shear, high-crystallinity corner and a high-temperature, high-shear, moderate-crystallinity corner; neither region contains data, and they can test the parabolic temperature window and square-root shear factor at the edge of the domain. Acquire a condition at moderate temperature and shear that pushes crystallinity toward the upper end of the public range to test whether the sine factor turns as predicted by its algebraic form.
\end{ScientificReportBox}

\end{document}